\documentclass{article}

\usepackage[utf8]{inputenc}
\usepackage{spverbatim}
\usepackage{changepage}
\usepackage{euler}
\usepackage{float}
\usepackage{subcaption}
\usepackage{lscape}
\usepackage{booktabs} 
\usepackage{rotating}
\usepackage{afterpage}
\usepackage{array}
\usepackage{caption}
\usepackage{pifont}
\usepackage{multirow}
\usepackage{hyperref}
\usepackage[T1]{fontenc}
\usepackage{xcolor}
\usepackage[linesnumbered,ruled,vlined]{algorithm2e}
\usepackage{hyperref} 
\usepackage{ragged2e}

\newcommand{\newauthor}[1]{\subsection*{#1}}
\newcommand{\affiliation}[1]{\small #1\\[0.5em]}
\newcommand{\email}[1]{\href{mailto:#1}{#1}}
\newcommand{\orcid}[1]{\href{https://orcid.org/#1}{ORCID: #1}}
\raggedright
\justifying 

\title{Artificial Intelligence for Infectious Disease Prediction and Prevention: A Comprehensive Review}

\begin{document}

\maketitle
\centering
\newauthor{Selestine MELCHANE}
\affiliation{$^1$Laboratoire LITAN, École supérieure en Sciences et Technologies de l’Informatique et du Numérique, RN 75, Amizour 06300, Bejaia, Algérie\\ 
$^2$LIASD research Lab., University of Paris 8, France}
\email{melchane@estin.dz}\\
\orcid{0009-0006-7902-6263}

\newauthor{Youssef ELMIR}
\affiliation{$^1$Laboratoire LITAN, École supérieure en Sciences et Technologies de l’Informatique et du Numérique, RN 75, Amizour 06300, Bejaia, Algérie\\ 
$^3$SGRE-Lab, Bechar, Algeria}
\email{elmir@estin.dz}\\
\orcid{0000-0003-3499-507X}

\newauthor{Farid KACIMI}
\affiliation{$^1$Laboratoire LITAN, École supérieure en Sciences et Technologies de l’Informatique et du Numérique, RN 75, Amizour 06300, Bejaia, Algérie\\ 
$^4$Laboratoire LIMED, Faculté des Sciences Exactes, Université de Bejaia, Algeria}
\email{kacimi@estin.dz}\\
\orcid{0000-0001-5457-2096}

\newauthor{Larbi BOUBCHIR}
\affiliation{$^2$LIASD research Lab., University of Paris 8, France}
\email{larbi.boubchir@univ-paris8.fr}\\
\orcid{0000-0002-5668-6801}

\justifying 






\begin{abstract}
Artificial Intelligence (AI) and infectious diseases prediction have recently experienced a common development and advancement. Machine learning (ML) apparition, along with deep learning (DL) emergence, extended many approaches against diseases apparition and their spread. And despite their outstanding results in predicting infectious diseases, conflicts appeared regarding the types of data used and how they can be studied, analyzed, and exploited using various emerging methods. This has led to some ongoing discussions in the field. This research aims not only to provide an overview of what has been accomplished, but also to highlight the difficulties related to the types of data used, and the learning methods applied for each research objective. It categorizes these contributions into three areas: predictions using Public Health Data to prevent the spread of a transmissible disease within a region; predictions using Patients' Medical Data to detect whether a person is infected by a transmissible disease; and predictions using both Public and patient medical data to estimate the extent of disease spread in a population. The paper also critically assesses the potential of AI and outlines its limitations in infectious disease management.
\end{abstract}

\textbf{keywords:} Infectious Diseases, Artificial Intelligence, Machine Learning, Prediction, Detection

\section{Introduction} 
For many years, the world has experienced several tragic events, with the emergence of diseases being among the most devastating upheavals. Considered as life companions for several decades, they have caused an increasingly dangerous imbalance in life. Defined as ``a particular abnormal condition that negatively affects the structure or function of all or part of an organism " \cite{IOMC}, they are generally associated with emerging signs and symptoms. Regarding causes, external factors like pathogens and internal malfunctions can be the origin of different diseases, distributed into various types. These include airborne diseases, foodborne diseases, lifestyle diseases, non-communicable diseases, and infectious diseases. Infectious diseases, also called communicable diseases, are among the most dangerous illnesses that haven't stopped manifesting and developing. They can spread from person to person or from animal to person \cite{kumar2020prediction}. Infectious diseases are classified into endemic diseases, epidemic diseases, and pandemic diseases. Endemic diseases, the first category, are identified diseases in a given region but do not spread. The second, epidemic diseases, is characterized by its rapid and brutal spread within a given region. The third, pandemic diseases, are communicable diseases that spread across continents or even the entire world, leading to the contamination of an unimaginable number of people. The classification of diseases is still unsatisfying; it has become not enough to just identify them, but also crucial to neutralize and prevent their spread in real time. Consequently, several studies have made the prediction and prevention of infectious diseases their main objective. Computers have significantly contributed to scientific advancement, especially with the vision of Artificial Intelligence (AI) and its impact on technological growth. The field of Machine Learning (ML) has significantly impacted research and opened the door to what was once considered impossible. Machines equipped with intelligence have now become essential tools in the world of science. Many ML models have therefore been designed to predict infectious diseases and forecast them. Various detection and prediction models have been developed due to the diverse learning approaches available. Depending on whether supervised, unsupervised, or semi-supervised learning (SSL) is employed, or whether classification or regression models are applied based on the research objective, several techniques can be used to create predictive models. Examples include Support Vector Machines (SVM), Decision Tree (DT) algorithms, clustering algorithms, Naive Bayes (NB), Artificial Neural Networks (ANN), Deep Neural Networks (DNN) such as Convolutional Neural Networks (CNN), Long Short-Term Memory (LSTM) networks and Transformers models, as well as emerging techniques like Transfer Learning. Understanding the development of specific diseases and their behaviors is crucial and greatly aids in identifying the origins of outbreaks and taking timely initiatives to combat them. In this paper, some applications of AI techniques in prediction and prevention of infectious diseases are introduced. The main focus is to define the various research objectives being considered, while analyzing which type of data is used to achieve each objective and which type of learning method to employ. A definition of the various research categories that have been developed is introduced to ensure an effective approach for building real-time prediction models for forecasting communicable diseases, while maintaining a swift and efficient predictive process.

This paper follows a specific structure : Section \ref{sec: research methodology} presents the research methodology used to introduce and discuss related works, which are extended in Section \ref{sec:relatedwork}. In Section \ref{sec:discussion}, a critical analysis and discussion of the accomplished work is presented. Finally, Section \ref{sec:conclusion} concludes the paper by highlighting potential future achievements and discussing future accomplishments. 
\section{Research methodology}\label{sec: research methodology}
Before taking a vision into the concepts discussed in the literature, three crucial questions caught attention :
\begin{itemize}
\item What are the various types of data that have been applied and studied?
\item Which kind of learning is being used?
\item How are the learning models selected based on the data?
\end{itemize} 
\color{black}
\begin{figure}[ht]
    \centering
         \includegraphics[scale=.38]{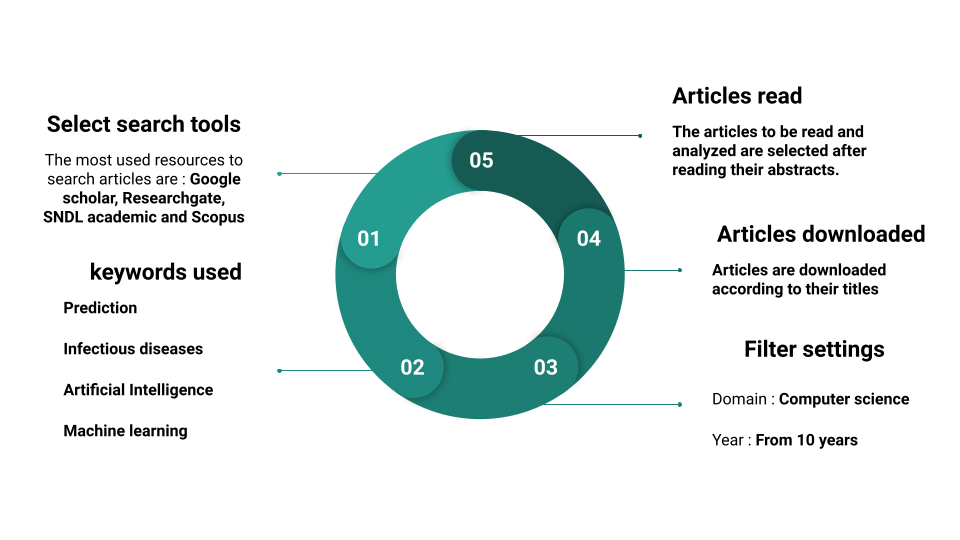}
         \caption{Research Methodology}\label{method}
\end{figure}
To address these issues, a comprehensive analysis and visualization of the completed work were conducted. For this, an extensive literature review was performed, focusing on relevant articles published since April 2022. This search was carried out using various academic search engines, such as : Google Scholar, ResearchGate, SNDL academic and Scopus, To obtain diverse and comprehensive electronic documentation on scientific research from various publishers. A combination of specific set of targeted keywords are also used, to enhance the precision of the search. In the electronic search phase, no restrictions were applied initially. Collecting early discoveries and foundational works on infectious diseases using ML is crucial for understanding the origins of predictions and the evolution of research. The second search was limited to articles published within the last ten years, using various keyword combinations. The key research areas and important topics studied are: infectious diseases, AI, ML and predictions. These articles were initially selected based on an assessment of their titles. Contributions were then further evaluated in the manual review phase after reading the abstracts. The criteria for selecting the articles and the resulting articles are illustrated in Figure \ref{method}.
Section \ref{sec:relatedwork} lists the contributions from the selected papers. The responses to the identified issues and the discussed limitations are covered in Section \ref{sec:discussion}.

\section{Background}\label{sec:relatedwork}
\begin{figure}[p]
    \centering
         \includegraphics[scale=.3]{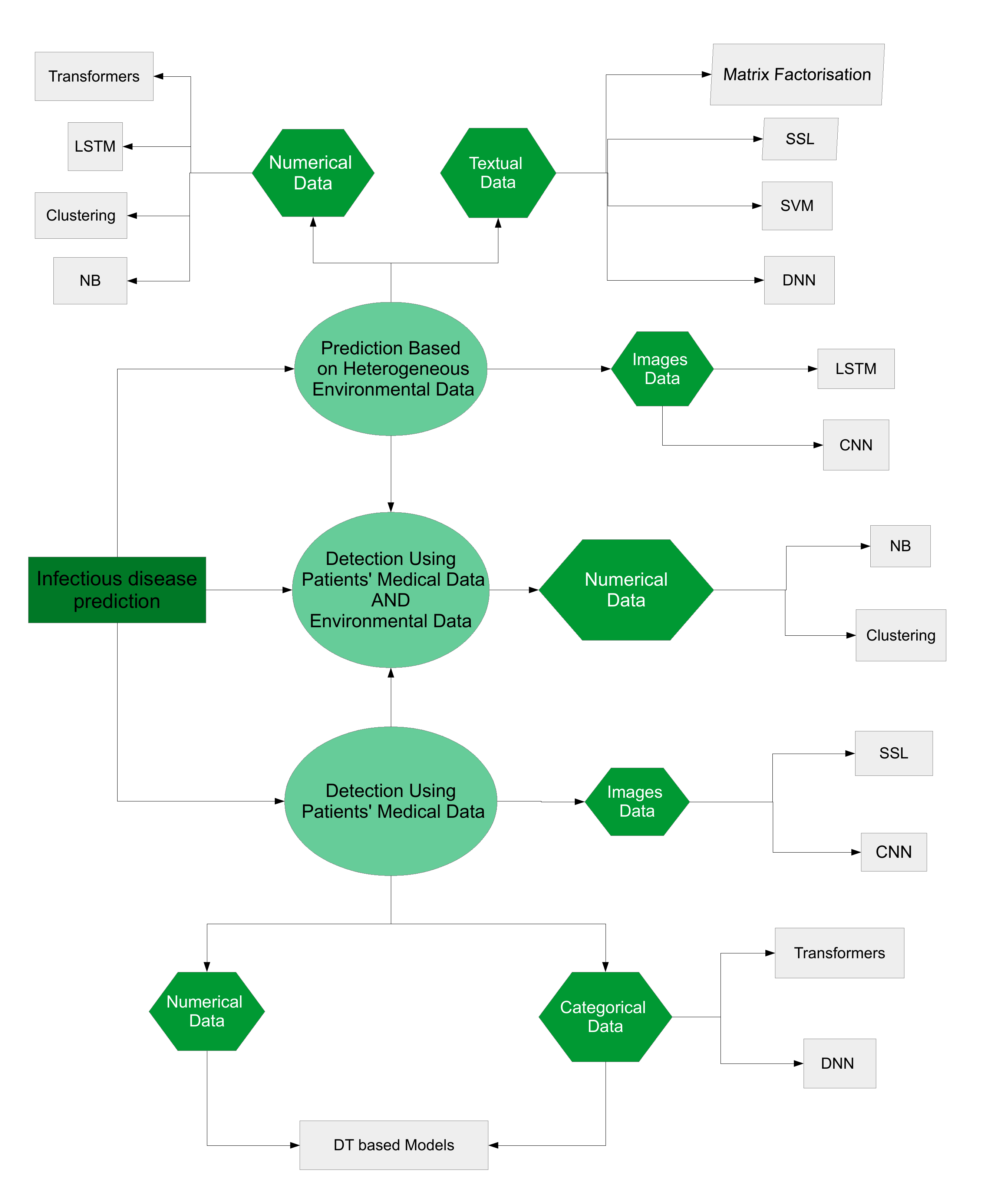}
         \caption{Diagram summarizing the techniques employed for each selected and studied dataset in the field of infectious disease prediction}\label{categorization}
\end{figure}

Human health is influenced by various life phenomena, and it depends not only on the individual themselves, but also on their surrounding environment. Consequently, individuals are called to address and adapt to these influences. Their ability to drive positive change through the development of innovative strategies and technologies offers hope in addressing various health crises. The appearance of the first pandemics that marked history, along with their impact on daily life and environmental changes, have pushed AI researchers around the world to develop various prediction and prevention systems. Three major categories of prediction research can be identified based on the research objectives and the data used: The first category focuses on detecting and predicting the spread of infectious diseases in specific locations. The second category aims to determine whether an individual may be infected by a communicable disease. While the third category combines the two first categories, addressing both the prediction of communicable diseases in patients and their spread within a population. The diverse methods and databases utilized in the detection of infectious diseases, as presented in this study, are illustrated in Figure \ref{categorization}.

\color{black}

Despite the distinctions that will be highlighted in the following sections between the three categories, all of them rely on the use of different AI techniques and share a common goal: preventing and controlling the spread of a disease.

\subsection{Prediction Based on Public Health Data}

Initial studies for preventing infectious diseases spread employed event-based and indicator-based surveillance methods \cite{christaki2015new}. However, their limitations in providing real-time predictions have directed research to the adoption of new technologies. Predictions based on Public Health Data enable a diverse range of technological applications. Consequently, datasets used to study disease's spread are not conform, and the methods applied are not the same for each data. The analysis of data collected from various sources such as social networks, news, mobile phones as well as environmental changes through geospatial images and Epidemiological data, has proven to be highly effective in tracking human displacements and movements, which helps in assessing pandemic transmissions. Studies have increasingly relied on the integration of multiple data types, including Numerical data, Textual data, and Image data.

\subsubsection{Numerical Data : Time-series Epidemiological and locational Data} In predicting contagious diseases, research databases are often presented in numerical formats. These data types frequently include location-related and epidemiological information. The most available data comprises information about the number of deaths, number of contaminated and number of recovered in regions.  Epidemiological data includes data on population exposure levels, which are essential for risk assessment. The most used algorithms for this type of data are Naive Bayes (NB), Clustering Algorithms, Long Short-Term Memory Based Models and Transformer Based Models.

\paragraph{Naive Bayes algorithms.}
Naïve Bayes is a simple yet robust algorithm for predicting outcomes. In machine learning, the goal is often to select the best hypothesis based on the given data. Naïve Bayes applies Bayes' Theorem, which offers a method for calculating the probability of a hypothesis using prior knowledge \cite{tiwari2022pandemic}. Tiwari et al. \cite{tiwari2022pandemic} involved the use of NB along with SVM, and Linear Regression (LR), to predict the trend of Covid-19 pandemic over the world while minimizing Mean Absolute Error (MAE) and Mean Squared Error (MSE) (Table \ref{Table13_NB}). The algorithms were applied to a real-time series dataset containing the global record of confirmed, recovered, deaths, and active cases of Covid-19 outbreak. Before the implementation phase, dataset pre-processing is also done for getting the effective results. During the fourth stage (Data collection, Data preprocessing, model training and model evaluation), the data is split into two subsets: the training set and the testing set, where 42 \% portion of the data is selected for testing predictions. The NB algorithm proved its effectiveness compared to other tested techniques with an MAE of 488806.7492 and MSE of 400919367451.7439. Despite its advantages, NB only works well with distinct and informative features \cite{raj2024classify}. Because it treats all features equally and presumes they are conditionally independent, its performance may suffer if there are noisy or irrelevant features \cite{raj2024classify}. 

\paragraph{Clustering Algorithms.} 
 Ravi et al. \cite{ravi2023novel} proposed a novel ML approach to track COVID-19 contact details that utilizes the DBSCAN algorithm, recognized as one of the most effective clustering algorithms. This approach incorporates time-series location data and prediction techniques to enhance tracking accuracy. The authors have proposed an innovative approach to prevent the spread of new infections in densely populated areas. DBSCAN is used as a clustering algorithm to locate infected individuals and their close contacts, in order to stop the transmission of the virus (Table \ref{Table13_NB}). In the study of Gupta et al. \cite{gupta2021clustering}, Two different clustering techniques, density-based clustering and partitioning-based clustering, were used to analyze COVID-19 infection cases. A comparative analysis was conducted between the DBSCAN and K-means algorithms, with DBSCAN showing better performance for clustering tasks. Although using time series locational data can provide valuable information about the movement patterns and interactions of individuals over time, DBSCAN has some disadvantages. Including high computational complexity and the need for careful selection of clustering parameters to ensure reliable results \cite{ravi2022critical}. Additionally, it does not work well with data of different densities and is not appropriate for high-dimensional data.

\begin{table}[!ht]
  \tiny
  \centering
  \caption{Performance evaluation of cited contributions in Time-series Epidemiological and locational Data observations with NB and Clustering algorithms} 
  \label{Table13_NB}
  \begin{tabular}{m{1cm} m{2.5cm} m{2cm} m{1.5cm} m{1.2cm} m{0.9cm} m{0.9cm}}
    \toprule
    \multirow{2}{*}{\textbf{Ref}} & \multirow{2}{*}{\textbf{Title}} & \multicolumn{2}{l}{\textbf{Evaluation Metrics}} \\ \cline{3-7}
    && \textbf{Dataset} & \textbf{Method} & \textbf{MAE (\%)} & \textbf{MSE (\%)} & \textbf{RMSE (\%)} \\
    \midrule
    \multirow{8}{*}{\begin{minipage}
    {1.5cm}\cite{tiwari2022pandemic} (2022)\end{minipage}} & \multirow{8}{*}{\begin{minipage}{2.5cm}Pandemic coronavirus disease (Covid-19): World effects analysis and prediction using machine-learning techniques\end{minipage}} \\
    && \multirow{1}{*}{{\begin{minipage}{1.5cm}real-time series dataset that holds the global record of confirmed, recovered, deaths\end{minipage}}} & {\begin{minipage}{1.5cm}NBN\end{minipage}} & 488806.74 & 400919367451.74 & {}\\
    && {} & {\begin{minipage}{1.5cm}SVM\end{minipage}} & 718150.13 & 565545811024.16 & {}\\
    && {} & {\begin{minipage}{1.5cm}LR\end{minipage}} & 648733.09 & 913583889578.49 & {}\\
    &&& & \\
    &&& & \\
    &&& & \\
    &&& & \\
    \cline{3-7}
     \multirow{10}{*}{\begin{minipage}
    {1.5cm}\cite{ravi2023novel} (2023)\end{minipage}} & \multirow{10}{*}{\begin{minipage}{2.5cm}A Novel Machine Learning Framework for Tracing Covid Contact Details by Using Time Series Locational data \& Prediction Techniques\end{minipage}} \\
    &&& & \\
    &&& & \\
    && \multirow{1}{*}{{\begin{minipage}{1.5cm}time series locational data\end{minipage}}} & {\begin{minipage}{1.5cm}MLDBSCAN\end{minipage}} & \multicolumn{3}{c}{\parbox{3cm}{\centering No evaluation metric provided. The proposed system helps in indicating to the user their respective output clusters and their contacts.}} \\
    &&& & \\
    &&& & \\
       \bottomrule
  \end{tabular}
\end{table}

\paragraph{Long short-term memory based models.}
Since LSTM based models are specialized in the exploration of times series data, they have the potential applications in the field of public health for forecasting epidemic cases, deaths, and recoveries. Some authors like Masum et al. \cite{masum2020covid} have produced a sustainable prognostic method of COVID-19 outbreak in Bangladesh using the DL models. The article presents a forecast on the counting number of infectious cases in Bangladesh from May 15th, 2020 until June 15th, 2020 (30 days). The LSTM network is used to predict the upcoming per day confirmed, death, and recovered cases in Bangladesh on patient data taken from the Institute of Epidemiology Disease Control \& Research (IEDCR) healthcare (Table \ref{Table13_LSTM}). The authors have also made a comparative analysis by the RMSE rate among the LSTM, Random Forest (RF) regression, and Support Vector Regression (SVR) models. Where LSTM proved higher performances on time series analysis. Zhou et al. \cite{zhou2023improved} have also presented a novel approach to forecasting COVID-19 using DL models. The proposed LSTM-based DL model is considered among the most advanced models to forecast time series data. They can take nonlinear factors into account and have the potential to provide more accurate predictions of COVID-19 cases, deaths, and recoveries. The proposed methodology in the paper involves constructing a prediction model of emergency material demand based on the infectious disease model. The model uses a time-varying demand and LSTM sequential decision model to provide a scientific and effective prediction method for actual emergency rescue work (Table \ref{Table13_LSTM} : MAE and MSE for death cases in United States). The approach combines traditional infectious disease prediction methods and DL prediction techniques. The proposed model also includes different countries' migration data, which helps to retrieve other characteristics related to the epidemic and accurately build the model.  Rakhshan et al. \cite{rakhshan2023global} for their part, have presented a combined approach for modeling and forecasting COVID-19, which can aid in determining interventions and predicting future growth patterns. The used dataset is sourced from the World Health Organization (WHO) and includes daily COVID-19 reports and global geographical distributions. The authors used this dataset to examine data from different countries, select targeted countries for their study, and collect COVID-19 data for analysis. In the authors' approach, dynamic epidemic models and ML methods work together to develop a package for predicting COVID-19. The authors used a dynamic model, along with five different ML models, to process the training and testing data. The dynamic model is a time-dependent compartmental model that captures fluctuations in the number of susceptible, infectious, and confirmed cases with controlled infectivity. The ML models, including LSTM, Multilayer perceptron (MLP), Adaptive neuro fuzzy inference system (ANFIS), General regression neural network (GRNN), and Radial basis function (RBF), are used to compare whether the classic dynamic model means would be best suited for predicting COVID-19 or the selected modern ML methods (Table \ref{Table13_LSTM}). And some metrics, including Root Mean Squared Error (RMSE), Relative Squared Error (RSE), and Accuracy, are used to evaluate the presented models. Another novel RNN-based model has been developed by Mu{\~n}oz-Organero et al. \cite{munoz2023new} to predict COVID-19 incidence in Madrid by integrating mobility data from a bike-sharing service. The model combines an LSTM-based RNN alongside mobility data to improve prediction accuracy. The analysis utilizes weekly COVID-19 case counts per district in Madrid and the number of bike rides recorded by the city’s bike-sharing service, BiciMAD. The bike-sharing data serves to estimate human mobility patterns between districts, while The LSTM RNN captures temporal patterns in the data. The proposed model achieves an RMSE of 0.0205 (Table \ref{Table13_LSTM}), outperforming the baseline model, which has an RMSE of 0.02296. This represents an 11.7\% improvement in prediction accuracy compared to the baseline model that excludes mobility data.

\begin{table}[!ht]
  \tiny
  \centering
  \caption{Performance evaluation of cited contributions in Time-series Epidemiological and locational Data observations using LSTM based models} 
  \label{Table13_LSTM}
  \begin{tabular}{m{1cm} m{2.5cm} m{2cm} m{1.5cm} m{1.2cm} m{0.9cm} m{0.9cm}}
    \toprule
    \multirow{2}{*}{\textbf{Ref}} & \multirow{2}{*}{\textbf{Title}} & \multicolumn{2}{l}{\textbf{Evaluation Metrics}} \\ \cline{3-7}
    && \textbf{Dataset} & \textbf{Method} & \textbf{MAE (\%)} & \textbf{MSE (\%)} & \textbf{RMSE (\%)} \\
    \midrule
    \multirow{16}{*}{\begin{minipage}
    {1.5cm}\cite{masum2020covid} (2020)\end{minipage}} & \multirow{16}{*}{\begin{minipage}{1.5cm}COVID-19 in Bangladesh: a deeper outlook into the forecast with prediction of upcoming per day cases using time series\end{minipage}} & \\
    && \multirow{3}{*}{{\begin{minipage}{1.5cm}Confirmed\end{minipage}}} & LSTM & {} & {} & 65.83 \\
    &&& RFR & {} & {} & 184.21 \\
    &&& SVR & {} & {} & 166.15 \\
    \cline{4-7}
    \multirow{3}{*}{} & \\ 
    && \multirow{3}{*}{{\begin{minipage}{1.5cm}Death\end{minipage}}} & LSTM & {} & {} & 2.95 \\
    &&& RFR & {} & {} & 3.28 \\
    &&& SVR & {} & {} & 4.73 \\
    \cline{4-7}
    \multirow{3}{*}{} & \\ 
    && \multirow{3}{*}{{\begin{minipage}{1.5cm}Recovery\end{minipage}}} & LSTM & {} & {} & 163.21 \\
    &&& RFR & {} & {} & 170.15 \\
    &&& SVR & {} & {} & 215.08 \\
    &&& & \\
    &&& & \\
    \cline{3-7}
    \addlinespace
     \multirow{7}{*}{\begin{minipage}{1.5cm}\cite{zhou2023improved} (2023)\end{minipage}} & \multirow{7}{*}{\begin{minipage}{2.5cm}Improved LSTM-based deep learning model for COVID-19 prediction using optimized approach\end{minipage}} &
    &&& & \\
    && \multirow{1}{*}{{\begin{minipage}{1.5cm}Epidemiological data\end{minipage}}} & \multirow{1}{*}{\begin{minipage}{1.5cm}LSTM\end{minipage}} & 0.01962 & 0.00102 & {} \\ 
    &&& {} {\begin{minipage}{1.5cm}GRU\end{minipage}} & 0.00679 & 0.02788 & {} \\ 
    &&& {} {\begin{minipage}{1.5cm}Bi-LSTM\end{minipage}} & 0.00623 & 0.25110 & {} \\ 
    &&& {} {\begin{minipage}{1.5cm}Dense-LSTM\end{minipage}} & 0.00763 & 0.00016 & {} \\ 
    &&& & \\
    &&& & \\
    \cline{3-7}
    \addlinespace
     \multirow{14}{*}{\begin{minipage}{1.5cm}\cite{rakhshan2023global} (2023)\end{minipage}} & \multirow{14}{*}{\begin{minipage}{2.5cm}Global analysis and prediction scenario of infectious outbreaks by recurrent dynamic model and machine learning models: A case study on COVID-19\end{minipage}} &
    &&& & \\
    && \multirow{1}{*}{{\begin{minipage}{2cm}daily COVID-19 reports and global geographical distributions : Trained data\end{minipage}}} & \multirow{1}{*}{\begin{minipage}{1.5cm}GRNN\end{minipage}} & {} & {} & 0.03 \\ 
     && {} & RBF & {} & {} & 0.006 \\ 
     && {} & LSTM & {} & {} & 0.25 \\  
     && {} & MLP & {} & {} & 0.005 \\ 
     && {} & ANFIS & {} & {} & 0.005 \\ 
    \cline{4-7}
    &&& & \\
    && \multirow{1}{*}{{\begin{minipage}{2cm}daily COVID-19 reports and global geographical distributions : Tested data\end{minipage}}} & \multirow{1}{*}{\begin{minipage}{1.5cm}GRNN\end{minipage}} & {} & {} & 0.06 \\ 
     && {} & \multirow{1}{*}{\begin{minipage}{1.5cm}RBF\end{minipage}} & {} & {} & 0.011 \\ 
     && {} & \multirow{1}{*}{\begin{minipage}{1.5cm}LSTM\end{minipage}} & {} & {} & 0.02 \\  
     && {} & \multirow{1}{*}{\begin{minipage}{1.5cm}MLP\end{minipage}} & {} & {} & 0.02 \\ 
     && {} & \multirow{1}{*}{\begin{minipage}{1.5cm}ANFIS\end{minipage}} & {} & {} & 0.01 \\ 
    &&& & \\
    \cline{3-7}
    \multirow{10}{*}{\begin{minipage}
    {1.5cm}\cite{munoz2023new} (2023)\end{minipage}} & \multirow{10}{*}{\begin{minipage}{2.5cm}A new RNN based machine learning model to forecast COVID-19 incidence, enhanced by the use of mobility data from the bike-sharing service in Madrid\end{minipage}} \\
    &&& & \\
    && \multirow{1}{*}{{\begin{minipage}{1.5cm}weekly COVID-19 case counts per district in Madrid and the number of bike rides recorded by the city’s bike-sharing service, BiciMAD \end{minipage}}} & {\begin{minipage}{1.5cm}LSTM-based RNN\end{minipage}} & {} & {} & 0.0205 \\
    &&& & \\
    &&& & \\
    &&& & \\
    &&& & \\
    &&& & \\
    &&& & \\
    &&& & \\
    &&& & \\
       \bottomrule
  \end{tabular}
\end{table}

\paragraph{Transformer Based Models.}
Transformers are neural network models that replace the commonly used recurrent layers in encoder-decoder architectures with multi-head self-attention. By relying entirely on attention mechanisms, transformers effectively capture global dependencies in data sequences and allow for much greater parallelization \cite{vaswani2017attention}. Ming et al. \cite{ming2023hostnet} developed a computational tool, HostNet, to predict virus hosts using deep neural networks. HostNet integrates Transformer, CNN, and BiGRU models, and was tested on a benchmark dataset of 'Rabies lyssavirus' and an in-house 'Flavivirus' dataset. It outperforms existing methods in accuracy and F1 score, thanks to its enhanced representation modules. Transformers are highly effective for time series forecasting tasks. Another Transformer-based model was also developed by Li et al. \cite{li2021long} to predict the long-term spread of seasonal influenza. It includes a source selection module to merge data from various sources and capture spatial dependencies. The model was tested on datasets from the United States and Japan, which included weekly influenza statistics from different regions. Demonstrating superior long-term forecasting performance compared to traditional autoregressive and RNN-based models, achieving an RMSE of 0.52 for short-term predictions and 0.87 for long-term predictions on the Japan dataset. For the US-HSS dataset, the model achieved an RMSE of 0.54 for short-term predictions and 0.89 for long-term predictions (Table \ref{Table13_TRANSFORMER}). Due to the limitations of traditional epidemiological and ML models in forecasting the COVID-19 pandemic such as challenges with generalization, scalability, and the lack of sufficient surveillance data, Wang et al. \cite{wang2022predicting} have proposed a novel approach that combines epidemiological theories with Generative Adversarial Networks (GANs). Their model, T-SIRGAN, integrates the Susceptible Infectious Recovered (SIR) model to generate epidemiological simulation data, while GANs are employed for data augmentation. Transformers are then used to predict future trends. The study utilized COVID-19 data, including cumulative confirmed cases and deaths, from the Center for Systems Science and Engineering (CSSE) at Johns Hopkins University. The T-SIRGAN model outperformed other methods, demonstrating superior accuracy in predicting epidemic trends by integrating epidemiological simulations and GANs. Specifically, the model achieved the lowest RMSE of 0.0188 for predicting confirmed cases, and an RMSE of 0.0243 for predicting death cases, outperforming other models in both metrics (Table \ref{Table13_TRANSFORMER}). Some other challenges in infectious disease prediction are addressed, such as the variability in incubation periods and the progression dynamics of different diseases. Wang et al. \cite{wang2023oriented} introduces an Oriented Transformer (ORIT), which improves upon traditional Multiple Representation Fusion (MRF) methods by capturing multi-dimensional temporal relationships within disease case data. ORIT incorporates a Multi-head Oriented Attention Unit (MOAU), designed to learn correlations from various orientations within the time series data, enabling the model to capture complex patterns in infectious disease progression. Two real-world datasets were used for evaluation: the Hand, Foot, and Mouth Disease (HFMD) dataset with 49,677 records, and the Hepatitis B Virus (HBV) dataset with 48,359 records. After data preprocessing, the MOAU captures attention from different orientations of the time series, including the impact of diverse time steps, correlations between different time series, and the significance of temporal segments. A comparison with 21 other models showed that ORIT demonstrated superior performance, achieving an RMSE of 16.8450 on the HFMD dataset and 28.2686 on the HBV dataset (Table \ref{Table13_TRANSFORMER}).

\begin{table}[!ht]
  \tiny
  \centering
  \caption{Performance evaluation of cited contributions in Time-series Epidemiological and locational Data observations using Transformer based models} 
  \label{Table13_TRANSFORMER}
  \begin{tabular}{m{1cm} m{2.8cm} m{3cm} m{2.5cm} m{1.5cm}}
    \toprule
    \multirow{2}{*}{\textbf{Ref}} & \multirow{2}{*}{\textbf{Title}} & \multicolumn{2}{l}{\textbf{Evaluation Metrics}} \\ \cline{3-5}
    && \textbf{Dataset} & \textbf{Method} & \textbf{RMSE (\%)} \\
    \midrule
    \multirow{6}{*}{\begin{minipage}
    {1.5cm}\cite{li2021long} (2021)\end{minipage}} & \multirow{6}{*}{\begin{minipage}{2.8cm}Long-term prediction for temporal propagation of seasonal influenza using Transformer-based model\end{minipage}} & \\
    && \multirow{1}{*}{{\begin{minipage}{3cm}weekly influenza-like-illness statistics JAPAN\end{minipage}}} & Transformer Short-term & 0.52 \\
    &&& Transformer Long-term & 0.87 \\
    &&& & \\
    \cline{4-5}
    \multirow{3}{*}{} & \\ 
    && \multirow{1}{*}{{\begin{minipage}{3cm}weekly influenza activity levels for 10 HHS regions of the U.S US-HSS\end{minipage}}} & Transformer Short-term & 0.54 \\
    &&& Transformer Long-term & 0.89 \\
    &&& & \\
    \cline{3-5}
    \cline{3-5}
     \multirow{5}{*}{\begin{minipage}{1.5cm}\cite{wang2022predicting} (2022)\end{minipage}} & \multirow{5}{*}{\begin{minipage}{2.8cm}Predicting the epidemics trend of COVID-19 using epidemiological-based generative adversarial networks\end{minipage}} \\
    && \multirow{1}{*}{{\begin{minipage}{2cm}COVID-19 data, including cumulative confirmed cases \end{minipage}}} & \multirow{1}{*}{\begin{minipage}{1.5cm}Transformer T-SIRGAN\end{minipage}} & 0.0188 \\ 
    &&& & \\
    &&& & \\
    \cline{4-5}
    &&& & \\
    && \multirow{1}{*}{{\begin{minipage}{2cm}COVID-19 data, including cumulative deaths cases \end{minipage}}} & \multirow{1}{*}{\begin{minipage}{1.5cm}Transformer T-SIRGAN\end{minipage}} & 0.0243 \\ 
    &&& & \\
    &&& & \\
    &&& & \\
    \cline{3-5}
    \multirow{5}{*}{\begin{minipage}
    {1.5cm}\cite{wang2023oriented} (2023)\end{minipage}} & \multirow{5}{*}{\begin{minipage}{2.8cm}Oriented transformer for infectious disease case prediction\end{minipage}} \\
    && \multirow{1}{*}{{\begin{minipage}{1.5cm}HFMD Dataset \end{minipage}}} & {\begin{minipage}{1.5cm}Oriented Transformer (ORIT)\end{minipage}} & 16.8450 \\
    &&& & \\
    \cline{4-5}
    &&& & \\
    && \multirow{1}{*}{{\begin{minipage}{1.5cm}HBV Dataset \end{minipage}}} & {\begin{minipage}{1.5cm}Oriented Transformer (ORIT)\end{minipage}} & 28.2686 \\
    &&& & \\
       \bottomrule
  \end{tabular}
\end{table}

\begin{table}[!t]
    \centering
    \begin{minipage}{\textwidth}
    \centering
    \tiny
        \resizebox{0.85\textwidth}{!}{%
        \begin{tabular}{|c|c|c|c|c|c|c|}
            \hline
            ObservationDate & Province/State & Country/Region & Last Update & Confirmed & Deaths & Recovered \\
            \hline
            01/22/2020 & Anhui & Mainland China & 1/22/2020 17:00 & 1.0 & 0.0 & 0.0 \\
            01/22/2020 & Beijing & Mainland China & 1/22/2020 17:00 & 14.0 & 0.0 & 0.0 \\
            01/22/2020 & Chongqing & Mainland China & 1/22/2020 17:00 & 6.0 & 0.0 & 0.0 \\
            01/22/2020 & Fujian & Mainland China & 1/22/2020 17:00 & 1.0 & 0.0 & 0.0 \\
            01/22/2020 & Gansu & Mainland China & 1/22/2020 17:00 & 0.0 & 0.0 & 0.0 \\
            01/22/2020 & Guangdong & Mainland China & 1/22/2020 17:00 & 26.0 & 0.0 & 0.0 \\
            01/22/2020 & Guangxi & Mainland China & 1/22/2020 17:00 & 2.0 & 0.0 & 0.0 \\
            01/22/2020 & Guizhou & Mainland China & 1/22/2020 17:00 & 1.0 & 0.0 & 0.0 \\
            01/22/2020 & Hainan & Mainland China & 1/22/2020 17:00 & 4.0 & 0.0 & 0.0 \\
            01/22/2020 & Hebei & Mainland China & 1/22/2020 17:00 & 1.0 & 0.0 & 0.0 \\
            01/22/2020 & Heilongjiang & Mainland China & 1/22/2020 17:00 & 0.0 & 0.0 & 0.0 \\
            01/22/2020 & Henan & Mainland China & 1/22/2020 17:00 & 5.0 & 0.0 & 0.0 \\
            01/22/2020 & Hong Kong & Hong Kong & 1/22/2020 17:00 & 0.0 & 0.0 & 0.0 \\
            01/22/2020 & Hubei & Mainland China & 1/22/2020 17:00 & 444.0 & 17.0 & 28.0 \\
            01/22/2020 & Hunan & Mainland China & 1/22/2020 17:00 & 4.0 & 0.0 & 0.0 \\
            01/22/2020 & Inner Mongolia & Mainland China & 1/22/2020 17:00 & 0.0 & 0.0 & 0.0 \\
            01/22/2020 & Jiangsu & Mainland China & 1/22/2020 17:00 & 1.0 & 0.0 & 0.0 \\
            01/22/2020 & Jiangxi & Mainland China & 1/22/2020 17:00 & 2.0 & 0.0 & 0.0 \\
            01/22/2020 & Jilin & Mainland China & 1/22/2020 17:00 & 0.0 & 0.0 & 0.0 \\
            01/22/2020 & Liaoning & Mainland China & 1/22/2020 17:00 & 2.0 & 0.0 & 0.0 \\
            \hline
        \end{tabular}
        }
        \captionof{table}{Time series example for confirmed, deaths and recovered cases \cite{Dataset} \cite{tiwari2022pandemic}}
        \label{Time_series_data}
    \end{minipage}
    \hspace{0.02\textwidth}
    \begin{minipage}{\textwidth}
    \centering
    \tiny
            \resizebox{0.85\textwidth}{!}{%
            \begin{tabular}{|l|r|r|r|}
            \hline
             Country/Region & Confirmed & Active & Deaths \\
            \hline
            US                & 1.528.568 & 1.147.255 & 91.921 \\
            Russia            & 299.941   & 220.974   & 2837  \\
            Brazil            & 271.885   & 147.108   & 17.983 \\
            UK                & 250.138   & 213.617   & 35.422 \\
            Spain             & 232.037   & 204.259   & 27.778 \\
            Italy             & 226.699   & 65.129    & 32.169 \\
            France            & 180.933   & 90.230    & 28.025 \\
            Germany           & 177.778   & 14.016    & 8081  \\
            Turkey            & 151.615   & 34.521    & 4199  \\
            Iran             & 124.603   & 20.311    & 7119  \\
            India            & 106.475   & 60.864    & 3302  \\
            Peru             & 99.483    & 60.045    & 2914  \\
            Mainland China   & 82.963    & 88        & 4634  \\
            Canada           & 80.493    & 34.396    & 6028  \\
            Saudi Arabia     & 59.854    & 27.891    & 329    \\
            Belgium          & 55.791    & 31.996    & 9108  \\
            Mexico           & 54.346    & 11.355    & 5666  \\
            Chile            & 49.579    & 27.563    & 509    \\
            The Netherlands  & 44.449    & 38.548    & 5734  \\
            Pakistan         & 43.966    & 30.538    & 939    \\
            \hline
        \end{tabular}
        }
         \captionof{table}{Top 20 Covid-19 affected countries record (confirmed, active and deaths) collected from 22 January 2020 to 19 May 2020 \cite{tiwari2022pandemic}}\label{Time_series}
    \end{minipage}
\end{table}

\section*{Discussion}
Using time series data to predict infectious diseases involves analyzing the epidemiological growth and contact tracing of the illness (Tables \ref{Time_series_data}, \ref{Time_series}). Locational data, for example, can help in identifying potential contacts and understanding the spread of the virus within a specific area. By analyzing their temporal aspect, it may be possible to track the movements of infected individuals and identify individuals who may have come into proximity with them. This can aid in effective contact tracing and containment strategies. However, due to limited accurate data on COVID-19 records and locations, as well as inherent uncertainties, traditional methods have struggled to accurately predict the global impact of the pandemic \cite{tiwari2022pandemic}. Recent ML models have shown improved efficiency in forecasting infectious diseases. Naïve Bayes proved its effectiveness in handling uncertainty by estimating the probabilities of outcomes, making it useful for both predictive and diagnostic tasks \cite{medhekar2013heart}. Clustering-based machine learning techniques can also automate contact tracing, resulting in more accurate and efficient outcomes \cite{gupta2021clustering}. Recurrent Neural Networks (RNNs) have gained significant attention in the field of deep learning for their ability to model nonlinear relationships. However, traditional RNNs face vanishing gradient issues and failed with capturing long-term dependencies \cite{bengio1994learning}. Long Short-Term Memory (LSTM) networks and their variants have been applied to sequence modeling, addressing these challenges and achieving success in various applications \cite{gers2000learning}. Transformer models have further demonstrated superior performance in capturing long-range dependencies compared to RNNs \cite{vaswani2017attention}, as their self-attention mechanism reduces the signal transmission path within the network, removing the need for a recurrent structure \cite{zhou2021informer}.

\subsubsection{Textual Data : Social and News Data}
Textual data is becoming increasingly important among the various types of data used to predict transmissible diseases, as it enhances monitoring and prevention efforts. Known for their real-time acquisition, many approaches are developed using social and news data. Techniques such as: superviseed matrix factorization, SSL, SVM and DNN have showed promising results.

\paragraph{Matrix factorization.}
Chakraborty et al. \cite{chakraborty2016extracting} used a supervised matrix factorization method to extract features of each disease from news streams. For each study, independent words collected from news related to the diseases are modeled into a matrix and combined with structured time series data from different outbreaks. Matrix factorization is then applied to factorize the initial matrix into two other matrixes, each one contains latent features which are used to detect disease apparition (Table \ref{Table2}). The method of detection used in this study involved collecting each word in relation to outbreaks, which proved to be time-consuming and less accurate. Although the study paid some attention to word extraction, these words were considered non-dependent, which contradicts the common addiction where the appearance of one word can influence the appearance of another \cite{raj2024classify}.

\paragraph{Semi-supervised learning based models.}
The ability of using News data has involved other approaches using different ML techniques. Kim et al. \cite{kim2021infectious} employed articles and reports to predict infectious diseases that did not occur for six months in various countries, testing models based on SVM, SSL, and DNN. The number of articles related to each disease was calculated, and diseases were then labeled for each country based on whether they have appeared or not. Known as an ML technique that combines labeled and unlabeled data, SSL based models showed outstanding performance compared with SVM and DNN (Table \ref{Table2}). Because SSL makes good use of both labeled and unlabeled data, it has attracted a lot of interest. This is particularly crucial in practical applications when very little data is labeled \cite{bao2024robust}. However, a lot of unimportant or noisy elements in real-world raw data are frequently missed by SSL approaches. To enhance semi-supervised classification performance, it is crucial to choose pertinent neighbors and characteristics for every sample \cite{bao2024robust}. And despite the quality of the study conducted in \cite{kim2021infectious}, the experience in their proposed work was conducted during two distinct periods, which can lead to a contradiction due to the existence of seasonal diseases. 

\paragraph{Support vector machine models.}
Since the SVM based models can handle high-dimensional problems with limited training data \cite{roy2023support}, Kim et al. \cite{kim2019weekly} developed a prediction model using SVM based on an analysis of articles related to influenza pandemics and infectious diseases. The authors extracted several keywords that were closely related to influenza and used word2vec to determine which keywords were related to the keyword `influenza'. Then, SVM was applied to the extracted data to predict if the number of influenza patients would increase or decrease at a specific week. The prediction results using news text data with SVM achieved a mean accuracy of 86.7\% in forecasting whether the weekly influenza-like illness (ILI) patient ratio would increase or decrease, and an RMSE of 0.611\% in estimating the weekly ILI patient ratio (Table \ref{Table2}). Thapen et al. \cite{thapen2016defender} used novel data-analytics to detect and forecast epidemics while developing DEFENDER system : Detecting and Forecasting Epidemics Using Novel Data-Analytics for Enhanced Response. The system ensures three services : early warning detection, situational awareness and nowcasting of epidemics. The number of tweets matching each symptom captured on the online database Freebase, was tracked daily for each geographical area monitored. To distinguish between health-related and non-health-related tweets and articles, two classifiers were used: SVM and NB (Table \ref{Table2}). The areas of high tweet activity were located in a country or region using the DBSCAN algorithm. The number of cases from the current data is predicted by adjusting the observed symptom levels to the previously available clinical data containing week, disease, location and count. Although SVM outperforms many other systems, it has limitations with complex data due to the high computational cost of solving quadratic programming problems \cite{tanveer2024comprehensive}. Its performance also heavily depends on the choice of kernel functions and their parameters \cite{tanveer2024comprehensive}.

\paragraph{Deep neural network based models.}
Since the study of Thapen et al. \cite{thapen2016defender} considered only a limited number of symptoms, Serban et al. \cite{serban2019real} proposed an improvement called SENTINEL of the previous system (DEFENDER), that aims to explore a boarder range of symptoms and diseases. DNN based models (CNN, LSTM) were then selected to differentiate between health-related and non-health-related tweets (Table \ref{Table2}). In both studies \cite{thapen2016defender} and \cite{serban2019real}, the social media twitter was analyzed. Given the strong correlation between infectious diseases and Twitter data \cite{shin2016high}, Chae et al. \cite{chae2018predicting} presents a novel approach for predicting infectious diseases using deep learning models, specifically DNN and LSTM, combined with big data sources like Twitter mentions along with Naver search queries and weather data. The study addresses limitations of traditional models like autoregressive integrated moving average (ARIMA), by incorporating real-time data to predict the spread of diseases such as chickenpox, scarlet fever, and malaria. The DL models significantly outperformed traditional methods, with DNN improving prediction performance by 24\% and LSTM by 19\% for chickenpox. The main evaluation metric, RMSE, showed that these models better captured trends. The mean RMSE of the top 10 DNN models for chickenpox was 72.8215, while the top 10 LSTM models had a mean RMSE of 78.2850, particularly during rapid disease spread. Demonstrating their potential to enhance infectious disease forecasting systems. Despite Twitter's reputation for being used by credible individuals sharing accurate information, it is not widely used by numerous people. Consequently, the conclusions obtained are then restricted.  Drinkall et al. \cite{drinkall2022forecasting}, for their part, have introduced a novel approach that incorporates transformer-based language models into infectious disease modeling using Reddit posts. The analysis uses Reddit comments extracted via the Pushshift API, state-level epidemiological data, government response data, and Google’s COVID-19 Community Mobility Reports, which provide local movement data. In the feature identification process, sentence-level encoding, dimensionality reduction, and clustering (HDBSCAN) are applied to isolate predictive features from Reddit comments. For evaluation, the resulting features are compared to traditional datasets in both a threshold-classification task and a time-series forecasting task. In the threshold-classification task, a Random Forest model utilizing the extracted features achieved the highest accuracy across various prediction horizons. Particularly in identifying upward trend signals for extreme events, with an average performance score of 0.880. In the time-series forecasting task, the transformer model consistently outperformed Gaussian Process and Martingale models. Achieving the lowest Root Mean Square Error (RMSE) of 0.0284 when the extracted features were used as covariates (Table \ref{Table2}). The method clearly outperforms traditional models in predicting COVID-19 trends, particularly in regions with unreliable epidemiological data.

\begin{table}[!ht]
  \tiny
  \centering
  \caption{Performance evaluation of cited contributions in News Data observations with Matrix factorization, SSL SVM and DNN models} 
  \label{Table2}
  \begin{tabular}{m{0.8cm} m{2.5cm} m{1cm} m{1cm} m{0.9cm} m{0.6cm} m{0.9cm} m{0.6cm} m{0.9cm}}
    \toprule
    \multirow{2}{*}{\textbf{Ref}} & \multirow{2}{*}{\textbf{Title}} & \multicolumn{7}{l}{\textbf{Evaluation Metrics}} \\ \cline{3-9}
    &&  \textbf{Dataset} & \textbf{Method} & \textbf{Precision (\%)} & \textbf{Recall (\%)} & \textbf{Accuracy (\%)} & \textbf{ROC (\%)} & \textbf{F1 score (\%)} \\
    \midrule
    \multirow{10}{*}{\begin{minipage}{1.5cm}\cite{chakraborty2016extracting} (2016)\end{minipage}} & \multirow{10}{*}{\begin{minipage}{2.5cm}Extracting signals from news streams for disease outbreak prediction\end{minipage}} & 
    \multirow{1}{*}{\begin{minipage}{1.5cm}Dengue\end{minipage}} & \multirow{10}{*}{\begin{minipage}{1.5cm}Supervised \newline Matrix \newline Factori-sation\end{minipage}} & 83.5 & 62.5 & {} & {} & {} \\ 
    \addlinespace
    &&\multirow{1}{*}{\begin{minipage}{1.5cm}Flu\end{minipage}} & {} & 79.3 & 58.5 & {} & {} & {} \\
    \addlinespace
    &&\multirow{1}{*}{\begin{minipage}{1.5cm}Malaria\end{minipage}} & {} & 81.2 & 68.5 & {} & {} & {} \\
    \addlinespace
    &&\multirow{1}{*}{\begin{minipage}{1.5cm}Diabetes\end{minipage}} & {} & 77.2 & 59.1 & {} & {} & {} \\
    \addlinespace
    &&\multirow{1}{*}{\begin{minipage}{1.5cm}TB\end{minipage}} & {} & 79.3 & 69.5 & {} & {} & {} \\
    &&& & \\
    \cline{3-9}
    \addlinespace
     \multirow{8}{*}{\begin{minipage}{1.5cm}\cite{kim2021infectious} (2021)\end{minipage}} & \multirow{8}{*}{\begin{minipage}{2.5cm}Infectious disease outbreak prediction using media articles with machine learning models\end{minipage}} &
     \multirow{1}{*}{} & \\ 
     && \multirow{1}{*}{\begin{minipage}{1.5cm}Articles \newline and \newline Reports\end{minipage}} & {\begin{minipage}{1.5cm}SSL\end{minipage}} & {} & {} & 83.3 & 79.1 & 83.2 \\ 
    \addlinespace
    &&\multirow{1}{*}{} & {\begin{minipage}{1.5cm}SVM\end{minipage}} & {} & {} & 73.2 & 65 & 76.9 \\
    \addlinespace
    &&\multirow{1}{*}{} & {\begin{minipage}{1.5cm}DNN\end{minipage}} & {} & {} & 80.6 & 74.6 & 81.9 \\
    &&& & \\
  \end{tabular}
  \begin{tabular}{m{0.8cm} m{3.5cm} m{2.5cm} m{1cm} m{0.9cm} m{0.9cm} m{0.6cm}}
    \toprule
    \multirow{2}{*}{\textbf{Ref}} & \multirow{2}{*}{\textbf{Title}} & \multicolumn{5}{l}{\textbf{Evaluation Metrics}} \\ \cline{3-7}
    &&  \textbf{Dataset} & \textbf{Method} & \textbf{Precision (\%)} & \textbf{Accuracy (\%)} & \textbf{RMSE (\%)} \\
    \midrule
    \multirow{5}{*}{\begin{minipage}{1.5cm}\cite{kim2019weekly} (2019)\end{minipage}} & \multirow{5}{*}{\begin{minipage}{3.5cm}Weekly ILI patient ratio change prediction using news articles with support vector machine\end{minipage}} &  
     \multirow{1}{*}{} & \\
     && {\begin{minipage}{2.5cm}Articles \newline influenza \newline infectious diseases\end{minipage}} & {\begin{minipage}{2.5cm}SVM\end{minipage}} & {} & 86.7 & 0.611 \\
     &&& & \\
     \cline{3-7}
     \multirow{6}{*}{\begin{minipage}
    {1.5cm}\cite{thapen2016defender} (2016)\end{minipage}} & \multirow{6}{*}{\begin{minipage}{3.5cm}DEFENDER: detecting and forecasting epidemics using novel data-analytics for enhanced response\end{minipage}} & 
    \multirow{10}{*}{} & \\ 
    && {\begin{minipage}{2.5cm}Social Medias (Twitter), News, Clinical Data\end{minipage}} & {\begin{minipage}{1.5cm}SVM, NB\end{minipage}} & 8.20 & {}\\ 
    &&& & \\
    &&& & \\
  \end{tabular}
  \begin{tabular}{m{1cm} m{3cm} m{2cm} m{1.5cm} m{1.8cm} m{0.9cm}}
    \toprule
    \multirow{2}{*}{\textbf{Ref}} & \multirow{2}{*}{\textbf{Title}} & \multicolumn{2}{l}{\textbf{Evaluation Metrics}} \\ \cline{3-6}
    && \textbf{Dataset} & \textbf{Method} & \textbf{RMSE} & \textbf{Accuracy (\%)} \\
    \midrule
     \multirow{6}{*}{\begin{minipage}{1.5cm}\cite{serban2019real} (2019)\end{minipage}} & \multirow{6}{*}{\begin{minipage}{3cm}Real-time processing of social media with SENTINEL: A syndromic surveillance system incorporating deep learning for health classification\end{minipage}} &
    \multirow{10}{*}{} & \\
    && {\begin{minipage}{1.5cm}News\end{minipage}} & \multirow{1}{*}{\begin{minipage}{1.5cm}CNN\end{minipage}} & {} & 93.9 \\ 
    && {\begin{minipage}{1.5cm}Twitter\end{minipage}} & {} & {} & 85.4 \\ 
    &&& & \\
    &&& & \\
    &&& & \\
    \cline{3-6}
     \multirow{5}{*}{\begin{minipage}{1.5cm}\cite{chae2018predicting} (2018)\end{minipage}} & \multirow{5}{*}{\begin{minipage}{3cm}Predicting infectious disease using deep learning and big data\end{minipage}} \\
    && \multirow{1}{*}{\begin{minipage}{1.5cm}Twitter mentions, Naver search queries and weather data\end{minipage}} & DNN & 72.8215 & {}  \\ 
    && {} & LSTM & 78.2850 & {} \\ 
    &&& & \\
    &&& & \\
    &&& & \\
    \cline{3-6}
    \multirow{6}{*}{\begin{minipage}{1.5cm}\cite{drinkall2022forecasting} (2022)\end{minipage}} & \multirow{6}{*}{\begin{minipage}{2.8cm}Forecasting COVID-19 caseloads using unsupervised embedding clusters of social media posts\end{minipage}} \\
    && \multirow{1}{*}{{\begin{minipage}{1.5cm}Reddit comments, state-level epidemiological data, government response data and Google’s COVID-19 Community Mobility Reports\end{minipage}}} & \multirow{1}{*}{\begin{minipage}{2.5cm}Transformer\end{minipage}} & {0.0284} \\ 
    &&& & \\
    &&& & \\
    &&& & \\
    &&& & \\
    &&& & \\
    &&& & \\
    &&& & \\
    &&& & \\
       \bottomrule
  \end{tabular}
\end{table}

\section*{Discussion}
\begin{figure}[ht]
    \centering
    \centering
         \includegraphics[height=8cm, width=13.5cm]{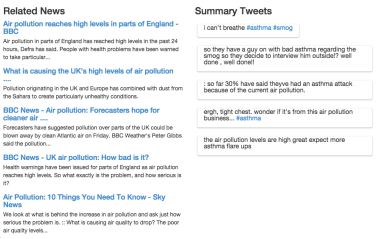}
         \caption{Situational Awareness screen showing asthma articles and Tweets in London from April 2014 \cite{thapen2016defender}}\label{Tweets}
\end{figure}

Some disease surveillance systems scan news articles from global sources like Google News and social media platforms such as Twitter (Figure \ref{Tweets}). They filter and classify these articles based on the type of epidemic, location, and news source. However, a major limitation of these systems is that they primarily focus on collecting disease-related information from various sources and compiling it for information dissemination or surveillance purposes \cite{chakraborty2016extracting}. A more precise and refined application of ML models is crucial for achieving optimal control over predictive systems. This ensures higher accuracy and effectiveness in decision-making processes. SSL based models, SVM based models along with DNN models showed high effectiveness in accurately distinguishing between health-related and non-health-related articles and tweets.

\subsubsection{Image Data : Geospatial Images}
Geospatial images, such as satellite images, are known as one of the most powerful and important tools for monitoring the earth \cite{sat-img}. They track the physical environment (water, air, land, vegetation) and the changing human footprint across the globe. And some DL algorithms specialized in image processing have given specific interest to explore Geospatial images to detect different symptoms related with a disease. 

\paragraph{Convolutional Neural Network based models.}
Regarding the importance of natural and environmental details, some authors ( Li et al.) \cite{li2023geoimagenet} started on the creation of a multi‑source natural feature benchmark dataset called GeoImageNet for GeoAI and supervised ML. The dataset was created by combining color imagery and Digital Elevation Model (DEM) data. GeoImageNet contains location information for each image scene, making geographic validation and training data expansion easy to achieve. The multi-source dataset empowers the machine to gain more geospatial intelligence and automation, resulting in higher prediction accuracy than commonly used single data sources. The authors have evaluated the dataset using two popular and representative object detection models, Faster-RCNN and Retina-Net, and its validity was proved for aiding a GeoAI model to achieve convergence and satisfactory detection performance (Table \ref{Table4}). CNN-based models excel at identifying important characteristics and successfully completing classification or prediction tasks \cite{li2023geoimagenet}. Thus, GeoImageNet has demonstrated its ability to support research on a wide range of environmental and health issues, including tracking the spread of infectious diseases.

\paragraph{LSTM based models.}
In order to examine the real world environment, Lee et al. \cite{lee2024convolutional} tried to develop a prediction model for the number of influenza patients at the national level using satellite images (Table \ref{Table4}). The authors developed a convolutional LSTM-LSTM neural network model, which demonstrated a strong correlation between the predicted and actual numbers of influenza patients, with an average MAE of 5.9010 per million population. Their study highlights the potential of using satellite image data as a valuable resource for predicting influenza incidence, which could facilitate timely national interventions. While the use of satellite images marks significant progress in real-time data acquisition, extracting and analyzing these images can be costly and requires specialized expertise \cite{moukheiber2024multimodal}. Additionally, satellite images are limited in their effectiveness for indoor localization due to restricted accessibility in indoor environments.

\begin{table}[!ht]
  \tiny
  \centering
  \caption{Performance evaluation of cited contributions in Geospatial Images data observations with LSTM and CNN based models} 
  \label{Table4}
  \begin{tabular}{m{1cm} m{2.5cm} m{2cm} m{1.5cm} m{1.5cm} m{1.9cm}}
    \toprule
    \multirow{2}{*}{\textbf{Ref}} & \multirow{2}{*}{\textbf{Title}} & \multicolumn{2}{l}{\textbf{Evaluation Metrics}} \\ 
    \cline{3-6}
    && \textbf{Dataset} & \textbf{Method} & \textbf{MAE} & \textbf{Precision (\%)} \\
    \midrule
    \addlinespace
    \multirow{6}{*}{\begin{minipage}{1.5cm}\cite{li2023geoimagenet} (2023)\end{minipage}} & \multirow{6}{*}{\begin{minipage}{2.5cm}GeoImageNet: a multi-source natural feature benchmark dataset for GeoAI and supervised machine learning\end{minipage}} &  
     \multirow{10}{*}{} & \\
     && {\begin{minipage}{3cm}single-source data\end{minipage}} &  \multirow{2}{*}{\begin{minipage}{1.5cm}Faster-RCNN and RetinaNet\end{minipage}} & {} & 50 \\
     \multirow{1}{*}{} & \\
     &&& & \\
     && {\begin{minipage}{1.5cm}GeoImageNet\end{minipage}} & {} & {} & 80 \\
     &&& & \\
     \cline{3-6}
    \addlinespace
     \multirow{8}{*}{\begin{minipage}{1.5cm}\cite{lee2024convolutional} (2024)\end{minipage}} & \multirow{8}{*}{\begin{minipage}{2.5cm}Convolutional LSTM–LSTM model for predicting the daily number of influenza patients in South Korea using satellite images\end{minipage}} &
    \multirow{1}{*}{} & \\
    && {\begin{minipage}{1.5cm}Sattelite Images\end{minipage}} & {\begin{minipage}{1.5cm}LSTM\end{minipage}} & 5.9010 per million population & {} \\ 
    &&& & \\
    &&& & \\
    &&& & \\
    &&& & \\
    &&& & \\
       \bottomrule
  \end{tabular}
\end{table}

\section*{Discussion}
Despite the use of various prediction models for infectious disease forecasting, most studies rely on local meteorological data, such as temperature, humidity, precipitation, and solar radiation. This limits predictions to specific regions and reduces their applicability at the national level \cite{lee2024convolutional}. Alternatively, satellite images provide a more comprehensive means of capturing weather patterns nationwide. These images are available through multiple channels, allowing the detection of key meteorological factors such as temperature, moisture, clouds, and precipitation. By proposing a national-level influenza prediction model based on satellite images and analyzing the relationship between influenza incidence and these meteorological factors, Lee et al. \cite{lee2024convolutional} present a model capable of forecasting influenza across the country. The integration of LSTM and CNN based models has, for instance, proven to be highly effective for forecasting infectious diseases using satellite imagery. However, due to the presence of certain areas that satellite images cannot capture, their use can be challenging \cite{yoneoka2024indirect}.

\subsection{Detection Using Patients' Medical Data}
The detection Using Patients' Medical Data category is focused in predicting whether a person is contaminated by the disease or not. Most researchers estimated that, more disease detection is quickly identified in a person, more the spread of disease will be controlled. This kind of detection is manifested through numerical data, categorical data and images data, using multiple techniques based on ML algorithms. 

\subsubsection{Numerical Data : Routine Blood Tests}
To evaluate the performance rate of ML applications in predicting diseases through routine blood tests, many collaborations test various ML models and algorithms. Peiffer-Smadja et al. \cite{peiffer2020machine} examined the exploration rate of ML in clinical microbiology. They concluded that : ANN, SVM, RF, LR, k-nearest neighbors (k-NN), and NB can be explored for targeting different diseases such as : bacterial infections, parasitic infections, viral infections, microorganism detection and diseases classification using diverse sources of data : microorganisms, microscopic-images and protein structure. Cabitza et al. \cite{cabitza2021development} for their part, developed, evaluated and validated ML models for COVID-19 detection using routine blood tests. ML models are developed following four steps : Imputation, Data normalization, Feature Selection and Classification. For the classification step, five different models (RF, NB, LR, SVM, k-NN) are developed for each kind of data: OSR patients, OSR dataset, covid-specific and CBC dataset. Models are therefore evaluated according to the Accuracy, Sensitivity, Specificity, Area under the curve (AUC) and External Validation. The Three best models extracted are : k-NN for COVID-19 specific dataset, RF and k-NN for CBC dataset (Table \ref{Table9}). Models based on DT Algorithms such as: DT, RF and Gradient Boosting, have showed most interest, particularly in exploring routine blood tests. DTs have the ability to explain ML-based diagnoses \cite{alves2021explaining}. They can break down complex data into more manageable parts, making them more interpretable compared to other algorithms in this category of prediction. 

\paragraph{Decision Tree based models.}
Among models based on DT, Random Forests (RF), a popular ML algorithm that aggregates the outputs of multiple decision trees to produce a unified result \cite{RF}. By aggregating the predictions of multiple trees, it reduces the risk of overfitting and improves the generalization ability of the model \cite{GRF}. While RF is not as easily interpretable as a single decision tree, it still provides insights into feature importance and evaluate the significance of various features, enabling the identification of the most influential ones for prediction. RF is capable of handling both categorical and continuous features \cite{GRF}. Brinati et al. \cite{brinati2020detection} focused on developing two ML models: RF Classifier and Three-Way RF Classifier (TWRF) using routine blood exams with demographic characteristics in order to detect COVID-19 infections. A DT Model is then interpreted to assist scientists in making decisions regarding the infections or not of COVID-19 (Table \ref{Table9}). For Banerjee et al. \cite{banerjee2020use}, They have guided their researches to predict if a person is SARS-CoV-2 positive or negative in the early stage of the disease from full blood counts. RF, glmnet (lasso-elastic-net regularized generalized linear) and ANN are used (Table \ref{Table9}). They indicated that RF and glmnet provided more information about important variables and clearly indicate how the decision was made.
For their iterative learning, boosting algorithms improved their importance in several studies. Kukar et al. \cite{kukar2021covid} started in performing COVID-19 diagnosis using Smart Blood Analytics (SBA) Algorithm and routine blood tests with the most popular ML tools, XGBoost to build the diagnostic models. The two most discriminating parameters were prothrombin and INR (International Normalized Ratio). For the evaluation of the model, ten-fold cross-validation is applied on independent testing data. XGBoost showed better results compared to other algorithms : SVM, RF and NN (Table \ref{Table9}). SBA technique is also employed in the study of Yang et al. \cite{yang2020routine}, where the authors aimed to predict an individual's SARS-CoV-2 infection status by employing Gradient Boosting Decision Tree (GBDT) using corporating patient demographic features such as age, sex, race and 27 routine laboratory tests (Table \ref{Table9}). Compared to LR, DT and RF, GBDT showed better results with 85.4 \% AUC, 76.1 \% sensitivity and 80.8 \% specificity. Given the highly accurate predictions from RF and GBoost, Yang et al. \cite{aljame2020ensemble} have called for the inclusion of both these models along with extremely randomized trees (ET) and LR models (Table \ref{Table9}). The authors have proposed a two-step learning approach for diagnosing COVID-19 using routine blood tests. The first step consists of making predictions using three different learning algorithms: ET, RF and LR. Resulting predictions are used in the second step as inputs of the prediction model XGBoost to establish the final predictions. The suggested model ERLX showed better results compared to previously proposed systems. However, the vulnerability in this proposed study is located in feature selection. Specifically, 18 features are selected to make the study according to the feature's importance appeared in older papers.
\newline
\begin{table}[!ht]
  \tiny
  \centering
  \caption{Performance evaluation of cited contributions in routine blood tests observations using DT based algorithm} 
  \label{Table9}
  \begin{tabular}{m{1cm} m{1.8cm} m{1.5cm} m{1.2cm} m{1cm} m{1cm} m{1cm} m{1.2cm}}
    \toprule
    \multirow{2}{*}{\textbf{Ref}} & \multirow{2}{*}{\textbf{Title}} & \multicolumn{2}{l}{\textbf{Evaluation Metrics}} \\ \cline{3-8}
    && \textbf{Dataset} & \textbf{Method} & \textbf{Accuracy (\%)} & \textbf{Sensitivity (\%)} & \textbf{Specificity (\%)} & \textbf{External-Validation (\%)} \\
    \midrule
    \multirow{10}{*}{\begin{minipage}
    {1.5cm}\cite{cabitza2021development} (2021)\end{minipage}} & \multirow{10}{*}{\begin{minipage}{1.8cm}Development, evaluation, and validation of machine learning models for COVID-19 detection based on routine blood tests\end{minipage}} \\
    && \multirow{1}{*}{{\begin{minipage}{1.5cm}routine blood tests Covid specific dataset\end{minipage}}} & {\begin{minipage}{1.5cm}k-NN\end{minipage}} & 78 & 74 & 81 & 94\\
    &&& & \\
    &&& & \\
    \cline{4-8}
    &&& & \\
    && \multirow{1}{*}{{\begin{minipage}{1.5cm}routine blood tests CBC dataset\end{minipage}}} & {\begin{minipage}{1.5cm}RF\end{minipage}} & 76 & 70 & 82 & 96\\
    &&& k-NN & 75 & 72 & 78 & 92\\ \\
    &&& & \\
    \cline{3-8}
    \addlinespace
    \multirow{10}{*}{\begin{minipage}
    {1.5cm}\cite{alves2021explaining} (2021)\end{minipage}} & \multirow{10}{*}{\begin{minipage}{1.8cm}Explaining machine learning based diagnosis of COVID-19 from routine blood tests with decision trees and criteria graphs\end{minipage}} \\
    && \multirow{1}{*}{{\begin{minipage}{1.5cm}Routine Blood Tests\end{minipage}}} & {\begin{minipage}{1.5cm}LR\end{minipage}} & 82 & 73 & 84 & {}\\
    &&& RF & 88 & 66 & 91 & {}\\ 
    &&& XGBoost & 87 & 60 & 91 & {}\\ 
    &&& SVM & 84 & 56 & 89 & {}\\ 
    &&& MLP & 85 & 42 & 92 & {}\\ 
    &&& ENSEMBLE & 88 & 67 & 91 & {}\\ 
    &&& & \\
    &&& & \\
  \end{tabular}
  \begin{tabular}{m{1cm} m{2.5cm} m{2cm} m{1.5cm} m{1cm} m{1cm} m{1cm}}
    \toprule
    \multirow{2}{*}{\textbf{Ref}} & \multirow{2}{*}{\textbf{Title}} & \multicolumn{2}{l}{\textbf{Evaluation Metrics}} \\ \cline{3-7}
    && \textbf{Dataset} & \textbf{Method} & \textbf{Accuracy (\%)} & \textbf{Sensitivity (\%)} & \textbf{Specificity (\%)} \\
    \midrule
    \multirow{8}{*}{\begin{minipage}
    {1.5cm}\cite{brinati2020detection} (2020)\end{minipage}} & \multirow{8}{*}{\begin{minipage}{2.5cm}Detection of COVID-19 infection from routine blood exams with machine learning: a feasibility study\end{minipage}}\\
    && \multirow{1}{*}{{\begin{minipage}{1.5cm}Routine Blood Tests\end{minipage}}} & {\begin{minipage}{1.5cm}DT\end{minipage}} & 70..78 & {} & {} \\
    &&& ET & 68..79 & {} & {} \\
    &&& k-NN & 66..76 & {} & {} \\
    &&& LR & 70..81 & {} & {} \\
    &&& NB & 64..81 & {} & {} \\
    &&& RF & 74..80 & {} & {} \\
    &&& SVM & 69..80 & {} & {} \\
    &&& TWRF & 83..89 & {} & {} \\
    &&& & \\
    \cline{3-7}
    \addlinespace
      \multirow{6}{*}{\begin{minipage}
    {1.5cm}\cite{banerjee2020use} (2020)\end{minipage}} & \multirow{6}{*}{\begin{minipage}{2.5cm}Use of machine learning and artificial intelligence to predict SARS-CoV-2 infection from full blood counts in a population\end{minipage}} \\
    && \multirow{1}{*}{{\begin{minipage}{1.5cm}Full Blood Counts\end{minipage}}} & {\begin{minipage}{1.5cm}RF\end{minipage}} & 82 & 60 & 88 \\
    &&& GLmnet & 81 & 65 & 81 \\
    &&& Ann & 87 & 43 & 91 \\
    &&& & \\
    &&& & \\
  \end{tabular}
  \begin{tabular}{m{1cm} m{2cm} m{1.5cm} m{1.5cm} m{1cm} m{1cm} m{0.5cm} m{1cm}}
    \toprule
    \multirow{2}{*}{\textbf{Ref}} & \multirow{2}{*}{\textbf{Title}} & \multicolumn{2}{l}{\textbf{Evaluation Metrics}} \\ \cline{3-8}
    && \textbf{Dataset} & \textbf{Method} & \textbf{Sensitivity (\%)} & \textbf{Specificity (\%)} & \textbf{AUC (\%)} & \textbf{Accuracy (\%)} \\
    \midrule
    \multirow{3}{*}{\begin{minipage}{1.5cm}\cite{kukar2021covid} (2021)\end{minipage}} & \multirow{1}{*}{\begin{minipage}{2cm}COVID-19 diagnosis by routine blood tests using machine learning\end{minipage}} \\
    && \multirow{1}{*}{{\begin{minipage}{1.5cm}Routine Blood Tests\end{minipage}}} & {\begin{minipage}{1.5cm}XGBoost ML Algorithm\end{minipage}} & 81.9 & 97.9 & 0.97 & {} \\
    &&& & \\
    \cline{3-8}
    \addlinespace
    \multirow{7}{*}{\begin{minipage}{1.5cm}\cite{yang2020routine} (2020)\end{minipage}} & \multirow{7}{*}{\begin{minipage}{2cm}Routine laboratory blood tests predict SARS-CoV-2 infection using machine learning\end{minipage}} \\
    && \multirow{1}{*}{{\begin{minipage}{1.5cm}Corporating patient demographic, Routine Laboratory Tests\end{minipage}}} & {\begin{minipage}{1.5cm}GBDT\end{minipage}} & 76.1 & 80.8 & 85.4 & {} \\
    &&& RF & 73.5 & 81.8 & 84.3 & {} \\
    &&& LR & 71.1 & 75.6 & 80.9 & {} \\
    &&& DT & 61.8 & 73.2 & 70.4 & {} \\
    &&& & \\
    &&& & \\
     \cline{3-8}
     \addlinespace
    \multirow{3}{*}{\begin{minipage}
    {1.5cm}\cite{aljame2020ensemble} (2020)\end{minipage}} & \multirow{3}{*}{\begin{minipage}{2cm}Ensemble learning model for diagnosing COVID-19 from routine blood tests\end{minipage}} \\
    && \multirow{1}{*}{{\begin{minipage}{1.5cm}Routine Blood Tests\end{minipage}}} & {\begin{minipage}{1.5cm}ERLX Ensemble learning model\end{minipage}} & 98,72 & 99,99 & 99,38 & 99,88 \\
    &&& & \\
       \bottomrule
  \end{tabular}
\end{table}

\section*{Discussion}
DT based models have demonstrated their effectiveness in detecting infectious diseases using routine blood data (Figure \ref{RB}). They provide deeper insights into the importance of various features and their significance. Models based on RF or GBoost, whether using parallel learning with RF or iterative learning with GBoost, allow for more accurate detection in the analysis of blood characteristics. The combination and hybridization of different decision tree algorithms further enhance accuracy and promise favorable results \cite{aljame2020ensemble}. Although the results may be promising, DT based models showed some limitations in depth selection. The choice of the top-level to derive the top consistent parameters can significantly impact the relevance of the model. Future research can address these limitations to develop more effective approaches. The use of blood data also presents confidentiality concerns with some significant missing values, making its application more difficult.
\begin{table}[t]
    \centering
    \tiny
\resizebox{0.85\textwidth}{!}{%
\begin{tabular}{|l|c|c|c|c|c|}
    \hline
    Parameter & Acronym & Unit of measure & COVID\-specific features & CBC features & Missing rate, \% \\
    \hline
    White blood cells & WBC & $10^9$/L & X & X & 2.4 \\
    Red blood cells   & RBC & $10^{12}$/L & X & X & 3.6 \\
    Hemoglobin        & HGB & g/dL & X & X & 2.4 \\
    Hematocrit        & HCT & \% & X & X & 2.4 \\
    Mean corpuscular volume & MCV & fL & X & X & 3.6 \\
    Mean corpuscular hemoglobin & MCH & pg/Cell & X & X & 3.6 \\
    Mean corpuscular hemoglobin concentration & MCHC & g Hb/dL & X & X & 2.4 \\
    Erythrocyte distribution width & RDW & CV\% & X & X & 3.7 \\
    Platelets         & PLT & $10^9$/L & X & X & 3.6 \\
    Mean platelet volume & MPV & fL & X & X & 5.9 \\
    Neutrophils count (\%) & NE & \% & X & X & 18.9 \\
    Lymphocytes count (\%) & LY & \% & X & X & 15.2 \\
    Monocytes count  (\%) & MO & \% & X & X & 15.2 \\
    Eosinophils count & (\%) EO & \% & X & X & 15.2 \\
    Basophils count (\%) & BA & \% & X & X & 15.2 \\
    Neutrophils count & NET & $10^9$/L & X & X & 15.2 \\
    Lymphocytes count & LYT & $10^9$/L & X & X & 15.2 \\
    Monocytes count & MOT & $10^9$/L & X & X & 18.9 \\
    Eosinophils count & EOT & $10^9$/L & X & X & 15.2 \\
    Basophils count   & BAT & $10^9$/L & X & X & 18.9 \\
    \hline
\end{tabular}
}
         \caption{Part of the Complete list of the analyzed features in the OSR dataset \cite{cabitza2021development}}\label{RB}
\end{table}

\subsubsection{Categorical Data : Clinical Data}
Clinical data are frequently used to track various diseases, and increased efficiency can be achieved by directly using factors in relation with biological experiments. This data can be integrated with different applying methods like DT based models, DNN based models, and Transformers. Indicating their efficiency in infectious disease detection. 

\paragraph{DT based models.}
As a parallel learning set model, RF algorithm has been the topic of various studies. In the study of Kumar et al. \cite{kumar2020prediction}, different ML models are adopted, among them : LR, RF, DT, MLP, SVM, k-NN, ANN, along with some other models, in order to predict chronic diseases (cardio vascular disease (CVD), chronic kidney disease (CKD), lung cancer) and infectious diseases (hepatitis and dengue serotypes) (Table \ref{Table1}). With Hepatitis Dataset analysis, RF exceed other algorithms with an accuracy of 90\%. The RF algorithm has therefore proved its effectiveness for extracting meaningful insights from data for predicting these kinds of infectious diseases. 

\paragraph{DNN based models.}
Some studies using clinical data take into consideration a few attributes and observations in diseases data, which can be limited, as it may indicate a disease other than the emerging one, since different diseases may share common symptoms. The reason why, Narmadha et al. \cite{narmadha2020intelligent} opted to select essential target proteins for cancer, Diabetes, Asthma and HPV viral infection by combining : color-graph algorithm (CG) and deep neural network model (DNN). The developed method CG-DNN takes biological data represented as a network interaction of protein-protein as input. Colors are assigned to proteins based on their connectivity with neighboring proteins. The three persistent colors are extracted, and equivalent proteins are considered as essential proteins to target. DNN model in this study is used to effectively learn five protein features, in order to improve the prediction accuracy of the essential protein (Table \ref{Table1}). Devi et al. \cite{devi2018mso} have for their part, demonstrated  high accuracy and less execution time in detecting DENV serotypes while using the MSO-MLP method (Table \ref{Table1}). MSO-MLP represents an incorporation of MLP and Multi-Swarm Optimization (MSO) which is a powerful metaheuristic algorithm building on the success of Particle Swarm Optimization (PSO), a population-based algorithm inspired by the collective behavior of bird flocks and fish schools. MSO advances this concept by incorporating multiple swarms instead of just one \cite{MSO}. In the MSO-MLP approach, multiple swarms of particles are used to optimize the weights and biases of the MLP \cite{devi2018mso}. Kumar et al. \cite{kumar2020prediction} have also studied the efficiency of MSO-MLP on Dengue Dataset (Table \ref{Table1}), where the MSO-MLP provides an accuracy of 86\%, which is better compared to other tested classifiers. 

\paragraph{Transformer Based Models.}
To design models with greater expressive power and computational efficiency, zhou et al. \cite{zhou2023tempo} chose to develop a method based on the transformer model. The attention architecture in transformers naturally enables better capture of long-range dependencies and supports large-scale parallel computation. The authors introduced TEMPO, a novel transformer-based model for predicting SARS-CoV-2 mutations using phylogenetic tree (PT)-based sequence sampling. The method generates evolutionary sequences from the viral phylogenetic tree and uses protein language pre-trained model ProtVec embeddings to represent these sequences, followed by a transformer model for site-specific mutation prediction. By incorporating phylogenetic information, TEMPO enhances prediction accuracy. The model is tested on a large dataset of over 7 million SARS-CoV-2 sequences, as well as influenza virus datasets (H1N1, H3N2, and H5N1), demonstrating its robustness. TEMPO outperforms traditional methods like SVM, logistic regression, and RNN-based models, showing a 1.1\% improvement in accuracy and a 6.5\% improvement in precision for SARS-CoV-2 mutation prediction.

\begin{table}[!t]
  \tiny
  \centering
  \caption{Performance evaluation of cited contributions in clinical data observations with DT, DNN and Transfomer models}
  \label{Table1}
  \begin{tabular}{m{1cm} m{4.5cm} m{2cm} m{1.5cm} m{1.5cm}}
    \toprule
    \multirow{2}{*}{\textbf{Ref}} & \multirow{2}{*}{\textbf{Title}}  & \multicolumn{3}{l}{\textbf{Evaluation Metrics}} \\ \cline{3-5}
    &&  \textbf{Dataset} & \textbf{Method} & \textbf{Accuracy (\%)}\\
    \midrule
    \multirow{23}{*}{\begin{minipage}{1.5cm}\cite{kumar2020prediction} (2020)\end{minipage}} & \multirow{23}{*}{\begin{minipage}{4.5cm}Prediction of chronic and infectious diseases using machine learning classifiers-A systematic approach\end{minipage}} & 
    \multirow{5}{*}{\begin{minipage}{2cm}Chronic Kidney disease\end{minipage}} & SVM & 64.5 \\ 
     && & C4.5 & 75.32 \\
     && & PSO-MLP & 68.31 \\
     && & DT & 72.67 \\
     && & ABC4.5 & 92.76 \\ \cline{4-5}
    \addlinespace
    &&\multirow{5}{*}{\begin{minipage}{2cm}Hepatitis Dataset\end{minipage}} & LR & 84 \\
    &&& RF & 90 \\
    &&& DT & 88 \\
    &&& C4.5 & 85 \\
    &&& MLP & 75 \\ \cline{4-5}
    \addlinespace
    &&\multirow{5}{*}{\begin{minipage}{2cm}CVD Dataset\end{minipage}} & RF & 80 \\
    &&& J48 & 85 \\
    &&& Hoeffding Tree & 86 \\
    &&& LMT & 85 \\
    &&& RT & 70 \\ \cline{4-5}
    \addlinespace
    &&\multirow{4}{*}{\begin{minipage}{2cm}Dengue Dataset\end{minipage}} & DT & 80 \\
    &&& ANN & 85 \\
    &&& MSO-MLP & 86 \\
    &&& PSO-ANN & 85 \\ \cline{4-5}
    \addlinespace
    &&\multirow{4}{*}{\begin{minipage}{2cm}Lung Cancer\end{minipage}} & SVM & 91.2 \\
    &&& k-NN & 83.2 \\
    &&& RF & 80.2 \\
    &&& ANN & 93.4 \\ \cline{3-5}
    \addlinespace
    \multirow{8}{*}{\begin{minipage}{1.5cm}\cite{narmadha2020intelligent} (2020)\end{minipage}} & \multirow{8}{*}{\begin{minipage}{4.5cm}An intelligent computer-aided approach for target protein prediction in infectious diseases\end{minipage}} &
    \multirow{1}{*}{} & \\
    && Biological Database \\
    &&& & \\
    && MIPS & \multirow{3}{*}{\begin{minipage}{1cm}CNN-DNN\end{minipage}}  & 91.87 \\
    && DIP & {}  & 93.12 \\
    && String DB & {} & 92.32 \\
    &&& & \\ \cline{3-5}
    \addlinespace
    \multirow{3}{*}{\begin{minipage}{1.5cm}\cite{devi2018mso} (2018)\end{minipage}} & \multirow{3}{*}{\begin{minipage}{4.5cm}MSO - MLP diagnostic approach for detecting DENV serotypes\end{minipage}} &
    \multirow{1}{*}{} & \\
    && Dengue Fever Dataset & \multirow{1}{*}{\begin{minipage}{1cm}MSO-MLP\end{minipage}}  & 85.18 \\
    &&& & \\ 
    \bottomrule
  \end{tabular}
\end{table}

\section*{Discussion}
Certain characteristics in clinical data, such as temperature, pulse, acute fever, vomiting, abdominal pain, body aches, cold symptoms, headache, weakness, fatigue, and rapid breathing, facilitate the identification of symptoms related to each disease \cite{kumar2020prediction}. Each infection has its own specific signs and symptoms, with common indicators including fever, diarrhea, fatigue, muscle aches, and coughing. Sequence data of the spike protein, along with its phylogenetic tree data, can also be utilized to select essential proteins \cite{narmadha2020intelligent} or predict virus mutations, such as amino acid changes in SARS-CoV-2 \cite{zhou2023tempo}. The application of ML and DL techniques has shown good accuracy in detecting infectious diseases using clinical data. Despite the productivity of clinical data and the highly effective detection capabilities of models based on DT, DNN and transformers, the similarity observed in the clinical symptoms of infectious diseases, frequently lead to disease identification issues \cite{ashraf2023early}. Furthermore, the use of clinical data is subject to privacy concerns, which makes data collection for studies quite challenging.  

\subsubsection{Images Data : Chest X-ray (CXR) and CT scan Images}
Another type of data used in detecting transmissible diseases in patients is image analysis. Specifically, chest X-ray images (CXR) and CT scan images are frequently collected in various studies focused on infectious disease detection using various models based on SSL technique and CNN model.

\paragraph{SSL based models.}
SSL addresses the limitations of supervised learning when dealing with datasets that include both labeled and unlabeled data. By using a small amount of labeled data along with a larger set of unlabeled data \cite{zhu2022introduction}. Sahoo et al. \cite{sahoo2021potential} have employed SSL approaches to detect COVID-19 cases accurately by analyzing digital chest X-rays and CT scans images. Their proposed algorithm COVIDCon applied on a small COVID-19 radiography dataset, attains 97.07\% average class prediction accuracy. When applied on large datasets, COVIDCon achieves an accuracy of 99.13\% (Table \ref{Table11}). The authors have therefore provided a fast, accurate, and reliable method for screening COVID-19 patients.

\paragraph{CNN based models.}
CNNs have proven to be a powerful class of models for comprehending the content of images, leading to significant advancements in image processing. CNNs are both efficient and effective in various pattern and image recognition applications, such as gesture recognition, face recognition, object classification, and generating scene descriptions \cite{sharma2018analysis}. Hussein et al. \cite{hussein2024auto} in their work, have discussed COVID-19 infections identification in chest X-rays by using Custom-CNN, a DL technique. The model achieved a classification accuracy of 98.19\% in distinguishing COVID-19, normal, and pneumonia samples (Table \ref{Table11}). Chest X-ray images were also employed in the study of Issahaku et al. \cite{issahaku2024multimodal} which focused on multimodality by integrating cough sound features. The Visual Geometry Group (VGG16) model was used for feature extraction and Faster R-CNN for COVID-19 detection. The study achieved an accuracy of 99.80\% (Table \ref{Table11}). As Transfer Learning approach enables the storage and use of knowledge acquired from pretrained models to address new problems \cite{hamida2021novel}, Some authors, Sadegji et al. \cite{sadeghi2024potential} introduced a novel dataset and proposed six different transfer learning models for slide-level analysis. Which was able to detect COVID-19 CT slides with an accuracy of more than 99\%. They have developed DL models to facilitate automated diagnosis of COVID-19 from CT scan records of patients. The authors also developed a novel 3D deep model (MASERes), for patient-level analysis, that achieved an accuracy of 100\% (Table \ref{Table11}). Enhancing the proposed models for practical use, especially in regions with limited medical infrastructure. Tan et al. \cite{tan2024self} also highlighted the importance of fine-tuning on small datasets to ensure the effectiveness of DL models. They proposed a method called Self-Supervised Learning with Self-Distillation for COVID-19 medical image classification (SSSD-COVID) (Table \ref{Table11}). CNN based models have therefore proven to be highly effective in detecting infectious diseases through the analysis of medical images.

\begin{table}[ht]
  \tiny
  \centering
  \caption{Performance evaluation of cited contributions in images data observations with SSL and CNN based models} 
  \label{Table11}
  \begin{tabular}{m{1cm} m{4cm} m{2cm} m{1.5cm} m{2cm}}
    \toprule
    \multirow{2}{*}{\textbf{Ref}} & \multirow{2}{*}{\textbf{Title}} &  \multicolumn{2}{l}{\textbf{Evaluation Metrics}} \\ \cline{3-5}
    &&  \textbf{Dataset} & \textbf{Method} & \textbf{Accuracy (\%)} \\
    \midrule
    \addlinespace
    \multirow{6}{*}{\begin{minipage}{1.5cm}\cite{sahoo2021potential} (2021)\end{minipage}} & \multirow{6}{*}{\begin{minipage}{4cm} Potential diagnosis of COVID‑19 from chest X‑ray and CT fndings using semi‑supervised learning
    \end{minipage}} &
    \multirow{7}{*}{} & \\ 
    && {\begin{minipage}{3cm}CT scans\end{minipage}} & {\begin{minipage}{1.5cm}SSL\end{minipage}} & 99.13 \\ 
    &&& & \\
    &&& & \\
    &&& & \\
    \cline{3-5}
     \multirow{7}{*}{\begin{minipage}{1.5cm}\cite{hussein2024auto} (2024)\end{minipage}} & \multirow{7}{*}{\begin{minipage}{4cm} Auto‐detection of the coronavirus disease by using deep convolutional neural networks and X‐ray photographs\end{minipage}} &
    \multirow{5}{*}{} & \\
    &&& & \\
    && {\begin{minipage}{3cm}chest X-rays\end{minipage}} & {\begin{minipage}{1.5cm}CNN\end{minipage}} & 98.19 \\ 
    &&& & \\
    &&& & \\
    &&& & \\
    \cline{3-5}
    \addlinespace
    \multirow{5}{*}{\begin{minipage}{1.5cm}\cite{issahaku2024multimodal} (2024)\end{minipage}} & \multirow{5}{*}{\begin{minipage}{4cm} Multimodal deep learning model for Covid-19 detection \end{minipage}} &  
     \multirow{1}{*}{} & \\
     && {\begin{minipage}{3cm}chest X-ray images \newline and cough sound\end{minipage}} &{\begin{minipage}{1.5cm}Vgg16, faster-RCNN\end{minipage}} & 99.80 \\
     \multirow{1}{*}{} & \\
     &&& & \\
     \cline{3-5}
     \addlinespace
    \multirow{5}{*}{\begin{minipage}{1.5cm}\cite{sadeghi2024potential} (2024)\end{minipage}} & \multirow{5}{*}{\begin{minipage}{4cm} Potential diagnostic application of a novel deep learning‐ based approach for COVID‐19 \end{minipage}} &  
     \multirow{1}{*}{} & \\
     && {\begin{minipage}{3cm}CT scans\end{minipage}} &{\begin{minipage}{1.5cm}Transfer Learning\end{minipage}} & 100 \\
     \multirow{1}{*}{} & \\
     &&& & \\
     \cline{3-5}
     \addlinespace
    \multirow{5}{*}{\begin{minipage}{1.5cm}\cite{tan2024self} (2024)\end{minipage}} & \multirow{5}{*}{\begin{minipage}{4cm} Self-supervised learning with self-distillation on COVID-19 medical image classification \end{minipage}} &  
     \multirow{1}{*}{} & \\
     && {\begin{minipage}{3cm}SARS-COV-CT \newline
     dataset\end{minipage}} &{\begin{minipage}{1.5cm}SSSD-COVID\end{minipage}} & 97.78 \\
     \multirow{1}{*}{} & \\
     &&& & \\
       \bottomrule
  \end{tabular}
\end{table}

\begin{figure}[t]
    \centering
    \begin{minipage}{\textwidth}
    \centering
         \includegraphics[height=4cm, width=13cm]{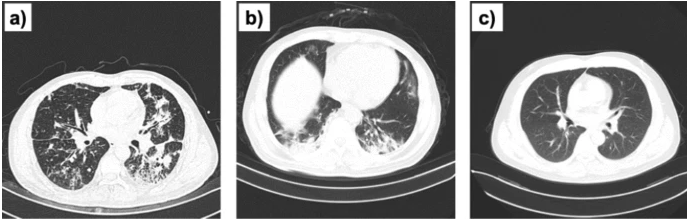}
         \caption{CT images taken from COVID-19 CT Scan dataset. Typical examples showing a) Common pneumonia (CP), b) COVID-19 (NCP), and c) normal CT scan image \cite{sahoo2021potential}}\label{CT}
    \end{minipage}
    \hspace{0.05\textwidth}
    \begin{minipage}{\textwidth}
    \centering
         \includegraphics[height=4cm, width=13cm]{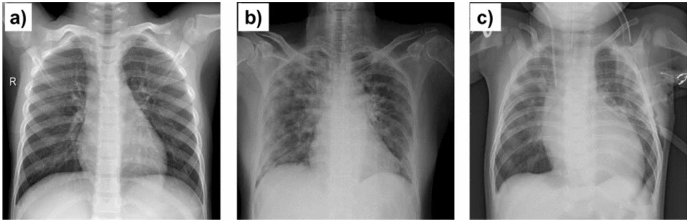}
         \caption{Sample chest X-rays taken from the COVID-19 Radiography dataset. a) Normal case, b) COVID-19 case showing bilateral ground-glass opacities with prominent peripheral, perihilar and basal distribution within a multilobar involvement, and c) viral pneumonia case with visible left basilar opacity \cite{sahoo2021potential}}\label{XCR}
    \end{minipage}
\end{figure}

\section*{Discussion}
In medical imaging, large unlabeled datasets are frequently available alongside smaller, high-quality labeled datasets. Consequently, SSL methods are a promising choice for automated medical image diagnosis (Figures \ref{CT}, \ref{XCR}). SSL, When paired with data augmentation and transfer learning, the approach can create powerful and more resilient models that require less training time \cite{sahoo2021potential}. CNN-based models also have the ability to recognize various image patterns, contributing to major advancements in image processing.

\subsection{Prediction Based on Public Health Data AND Patients' Medical Data}
Due to technological advancements and the implementation of various prediction and detection techniques, the monitoring of contagious diseases has achieved a high level of reliability in terms of exploring extensive data related to the spread of infectious diseases. Indeed, several studies have highlighted the importance of using both patients medical data along with Public Health Data to offer a complete and integrated approach to monitoring and predicting transmissible diseases, working for optimal effectiveness and real-time precision.

\subsubsection{Numerical Data : Environmental Data AND Patients' Data}
Many outbreaks are caused by environmental changes, highlighting the importance of addressing living conditions that contribute to the emergence of various infections. Zhang et al. \cite{zhang1995prevention} have for instance considered human behavior and its impacts on the environment. A set of actions and their associated impacts were used to develop a learning process using ML algorithms and advanced mathematical models. In order to denounce bad behavior and warn the environment of potential illnesses that may arise and deal with them, a collection of actions-impacts is carried out through the involvement of policymakers and international organizations. However, these actions-impacts rules considered non-permanent can lead to non-durable conclusions. New treatment methodologies have therefore taken place like Naive Bayes Network (NBN) and clustering algorithms, to predict the living conditions of a disease.
\paragraph{NB Network.}
In order to achieve disease prediction within a specific region, some studies attempted to use the NBN algorithm. Sood et al. \cite{sood2021intelligent} created an intelligent healthcare system for predicting and preventing dengue virus infection. They focused on changing environmental data using Individual health data. Once the health attributes of individuals and the environment have been analyzed using different technologies (sensors, mobile phones, etc.), these are passed to the Fog Computing system which performs data pre-processing. The NBN is used to classify individuals as IN (Potential infected) or UN (uninfected) in order to generate diagnoses, suggestions and alerts. GPS location is then used to identify the risks of spreading in each region (Table \ref{Table3}). The achievement of certain symptoms in this study is identified by the user himself, and technologies used may be less efficient in indoor locality. The results can therefore either conclude overfitting or underfitting. 

\paragraph{Clustering based algorithms.}
By using cluster analysis and factor analysis, Valiakos et al. \cite{valiakos2014use} combined environmental data and Human cases data with wild bird surveillance for predicting spatial distributions of West Nile Virus (WNV) in Greece. Data on 2010 and 2011 human cases are used for the statistical analysis and model building, and the 2012 cases are used for verification.  Cluster analysis was employed to cluster human cases and wild bird animals, while factor analysis was utilized to reduce the data and Principal Component Analysis (PCA) for extracting components. It was observed that altitude and distance from water were the two variables which clustered significantly in similar way humans and birds cases among the 37 variables under study. The obtained results lead for potential estimation of West Nile virus emerging (Table \ref{Table3}). The fact that, only resident WNV-seropositive wild birds were studied, even though samples from migratory birds tested positive, does not guarantee that the analysis conforms to the real environment. All cases tested positive have to be considered to greatly know the real illness origins.

\begin{table}[t]
  \tiny
  \centering
  \caption{Performance evaluation of cited contributions in Public Health AND Patients' Medical Data observations} 
  \label{Table3}
  \begin{tabular}{m{0.8cm} m{2cm} m{1.9cm} m{0.8cm} m{0.9cm} m{0.9cm} m{0.9cm} m{0.9cm}}
    \toprule
    \multirow{2}{*}{\textbf{Ref}} & \multirow{2}{*}{\textbf{Title}} &  \multicolumn{2}{l}{\textbf{Evaluation Metrics}} \\ \cline{3-8}
    &&  \textbf{Dataset} & \textbf{Method} & \textbf{Sensitivity (\%)} & \textbf{Specificity (\%)} & \textbf{Precision (\%)} & \textbf{Odds (\%)} \\
    \midrule
    \multirow{5}{*}{\begin{minipage}{1.5cm}\cite{sood2021intelligent} (2021)\end{minipage}} & \multirow{5}{*}{\begin{minipage}{2cm}An intelligent healthcare system for predicting and preventing dengue virus infection\end{minipage}} &
    \multirow{1}{*}{} & \\ 
    && {\begin{minipage}{1.9cm}Individual Health Data, Environmental Data\end{minipage}} & {\begin{minipage}{0.8cm}NBN, Hill Climbing (HL)\end{minipage}} & 94 & 95.1 & 89.8 & {} \\ 
    &&& & \\
    &&& & \\
    \cline{3-8}
     \multirow{8}{*}{\begin{minipage}{1.5cm}\cite{valiakos2014use} (2014)\end{minipage}} & \multirow{8}{*}{\begin{minipage}{2cm}Use of wild bird surveillance, human case data and GIS spatial analysis for predicting spatial distributions of West Nile virus in Greece\end{minipage}} &
    \multirow{1}{*}{} & \\
    && {\begin{minipage}{1.9cm}Wild Bird Animals, Human WNV cases Data (2010-2011 for training, 2012 for validation)\end{minipage}} & {\begin{minipage}{0.8cm}Cluster Analysis, Factor Analysis\end{minipage}} & {} & {} & {} & 95 \\ 
    &&& & \\
    &&& & \\
    &&& & \\
       \bottomrule
  \end{tabular}
\end{table}


\begin{table}[t]
  \tiny
  \centering
  \caption{Summary of Observations and Identified Limitations} 
  \label{Table12}
  \begin{tabular}{m{1.5cm} m{1.5cm} m{1.7cm} m{1.5cm} m{4cm}}
    \toprule
    \textbf{Prediction} & \textbf{Data} & \textbf{Confidentiality} & \textbf{Availability} & \textbf{Limitations} \\
    \midrule
    \multirow{14}{*}{\begin{minipage}{1.5cm}based on Public \newline Health Data\end{minipage}} & \\
    &&& & \\
    & \begin{minipage}{1.5cm}Time-series\end{minipage} & \textcolor{green}{\ding{55}} & \textcolor{green}{\ding{51}} & {\begin{minipage}{4cm}Although this type of data may have missing values, the variations in disease patterns across different studied periods also make it difficult to achieve accurate predictions and build generalized models. \end{minipage}} \\
    &&& & \\
    \cline{3-5}
    &&& & \\
    & \begin{minipage}{1.5cm}Social \newline News data \end{minipage} & \textcolor{green}{\ding{55}} & \textcolor{green}{\ding{51}} & {\begin{minipage}{4cm}The use of social media and news data is marked by uncertainty. Some tweets or news reports may not present the truth and are often inaccurate.\end{minipage}} \\
    &&& & \\
    \cline{3-5}
    &&& & \\
    & \begin{minipage}{1.5cm}Geospatial \newline images \end{minipage} & \textcolor{green}{\ding{55}} & \textcolor{green}{\ding{51}} & {\begin{minipage}{4cm} Although geospatial images are effective for capturing environmental conditions, their extraction and analysis can be expensive and require specialized expertise. Additionally, their applicability is limited in indoor spaces.\end{minipage}} \\
    &&& & \\
    \addlinespace
    \midrule
    \multirow{14}{*}{\begin{minipage}{1.5cm}Based on Patients'  medical data\end{minipage}} & \\
    &&& & \\
    & \begin{minipage}{1.5cm}Routine \newline Blood \newline Tests \end{minipage} & \textcolor{red}{\ding{51}} & \textcolor{red}{\ding{55}} & \multirow{10}{*}{\begin{minipage}{4cm}Medical data, including blood tests, clinical records, and images, are constrained by imbalanced data and a large number of missing values. Clinical data, like symptoms of fever and cough, can sometimes lead to confusion and uncertainty, as many infectious diseases share similar symptoms. \end{minipage}} \\
    &&& & \\
    \cline{3-4}
    &&& & \\
    & \begin{minipage}{1.5cm}Clinical \newline data \end{minipage} & \textcolor{red}{\ding{51}} & \textcolor{red}{\ding{55}} & \\
    &&& & \\
    \cline{3-4}
    &&& & \\
    & \begin{minipage}{1.5cm}CXR \newline CT images \end{minipage} & \textcolor{red}{\ding{51}} & \textcolor{red}{\ding{55}} & \\
    &&& & \\
    \addlinespace
    \midrule
    \multirow{6}{*}{\begin{minipage}{1.5cm}Based on Patients'  medical data, and Public Health data\end{minipage}} & \\
    &&& & \\
    & \begin{minipage}{1.5cm}Environmental \newline Patient \newline Data \end{minipage} & \textcolor{green}{\ding{55}} & \textcolor{green}{\ding{51}} & {\begin{minipage}{4cm}The challenge lies in the combined limitations of each type of data used. Some diseases have similar impacts on the environment, and it is difficult to distinguish which disease is spreading. \end{minipage}} \\
    &&& & \\
       \bottomrule
  \end{tabular}
\end{table}

\section{Discussion}\label{sec:discussion}
The application of AI, especially through the exploration of ML and DL techniques, has proven highly effective in detecting and predicting communicable diseases. The progression of ML has developed alongside the growth of diverse data types, which can be used for training, testing, and validating models under development. Although this research primarily provides an introduction and overview of the work done in predicting infectious diseases, and does not cover all the models and techniques that have emerged, its main focus is to define the research objectives, the types of data that can be used, and the learning methods that could be applied. For this purpose, the selection of a learning methodology depends on the availability of data and the specific research objectives to be analyzed and studied. Indeed, if the research aims to develop a model to detect and predict the spread of an infectious disease in specific regions, multiple types of data are required, including numerical, image, and text data. Several data sources are available for this purpose, such as epidemiological data, news reports, geospatial data, and social data. After the identification and processing of each kind of data related to prediction based on Public Health Data of infectious diseases, several approaches are used. With epidemiological data, LSTM and Transformers based models have yielded considerable outcomes in various studies and experiments behind NB and DBSCAN methods. Using news data, SSL, SVM along with DNN algorithms showed better results. While using geospatial images, LSTM and CNN architectures demonstrated their performances. Regarding the other detection category, using patients' medical data, the type of data used is more persistent, since the characteristics of clinical data, blood tests along with CXR images and CT scans, are generally stable with the same measurement and features and not frequently subject to change. All constructed models showed great performances across various evaluation metrics. Betters were DT based models: DT, RF and GBoost for routine blood tests. DNNs and Transformers based models have also performed with clinical data behind DT based algorithms. While for CXR images and CT scans, SSL and CNN based models excelled. However, individuals data suffer from constraints related to privacy. In the study of Kukar et al. \cite{kukar2021covid}, negative training data are randomly sampled to approximate the proportion of positif training group since there was a lack of data. And this way of preprocessing can lead to poor construction of predictive models. To this end, researchers have focused on studying both categories by exploring patient and environmental data to simultaneously identify individuals with communicable diseases and alert those who may be exposed to the risk of infection. This aids in identifying and locating the infectious disease. Learning methods based on clustering models and NB have demonstrated great effectiveness in processing this approach. Thereby enabling the collection of the most comprehensive information from each type of data used. Therefore, The used data plays an important role in prediction improvement. Bad or non-corresponding data often leads to overestimate or underestimate the outbreak rate for various reasons \cite{chakraborty2016extracting}. For example, when a user searches for information on a particular disease using the Google search engine, it doesn't necessarily mean they have that disease. The user might be researching one disease while actually having another, or they may not have any disease at all. This concerned all data related to social media, news data or environmental data. However, these types of data are known for their ease and real-time acquisition, in contrast to clinical data which include confidentiality and limited availability (Table \ref{Table12}. Furthermore, the learning algorithms are applied using different learning, testing and validation periods. So, each study is constrained to no longer be consistent for a period other than the already studied. Since infectious diseases are known for their seasonality and ability to rapidly mutate (Table \ref{Table12}. Allowing them to change their living conditions and their spread intensity. It is therefore recommended, as a future research perspectives to improve the prediction of infectious diseases, to first establish the research objectives and then carefully select the most appropriate learning approach and datasets. Attention should be focused on finding a method to integrate various types of data to collect comprehensive information for developing a general predictive model based on each evolution of the disease. This approach would address data gaps and provide accurate and well-supported insights for each studied period.

\section{Conclusion}\label{sec:conclusion}
In summary, the integration of AI into infectious disease prediction has shown promise through diverse ML models. Studies exploring various data sources, including social media, clinical records, and routine blood tests, highlight AI's adaptability. Despite significant progress, challenges persist. Privacy concerns, data quality issues, and models variability underscore the need for careful implementation. Ongoing challenges include the dynamic nature of infectious diseases, which can rapidly evolve and mutate. Additionally, biases in data can lead to skewed results. Looking forward, the combination of robust datasets and advanced AI techniques is essential for accurate outbreak predictions. Continuous refinement, addressing data biases, and selecting suitable models are crucial for unlocking the full potential of AI. In conclusion, While AI has made significant progress in forecasting infectious diseases, ongoing improvements and a nuanced approach are essential for achieving its full impact on global health.

\bibliographystyle{IEEEtran}
\bibliography{mybibfile}

\begin{thebibliography}{10}
\providecommand{\url}[1]{#1}
\csname url@samestyle\endcsname
\providecommand{\newblock}{\relax}
\providecommand{\bibinfo}[2]{#2}
\providecommand{\BIBentrySTDinterwordspacing}{\spaceskip=0pt\relax}
\providecommand{\BIBentryALTinterwordstretchfactor}{4}
\providecommand{\BIBentryALTinterwordspacing}{\spaceskip=\fontdimen2\font plus
\BIBentryALTinterwordstretchfactor\fontdimen3\font minus
  \fontdimen4\font\relax}
\providecommand{\BIBforeignlanguage}[2]{{%
\expandafter\ifx\csname l@#1\endcsname\relax
\typeout{** WARNING: IEEEtran.bst: No hyphenation pattern has been}%
\typeout{** loaded for the language `#1'. Using the pattern for}%
\typeout{** the default language instead.}%
\else
\language=\csname l@#1\endcsname
\fi
#2}}
\providecommand{\BIBdecl}{\relax}
\BIBdecl

\bibitem{IOMC}
``{Medical Reports \& Case Studies},''
  \url{https://www.iomcworld.org/medical-journals/diseases-online-journals-55099.html},
  2023.

\bibitem{kumar2020prediction}
N.~K. Kumar and K.~T. Sikamani, ``Prediction of chronic and infectious diseases
  using machine learning classifiers-a systematic approach,'' \emph{Int J
  Intell Eng Syst}, vol.~13, no.~4, pp. 11--20, 2020.

\bibitem{christaki2015new}
E.~Christaki, ``New technologies in predicting, preventing and controlling
  emerging infectious diseases,'' \emph{Virulence}, vol.~6, no.~6, pp.
  558--565, 2015.

\bibitem{tiwari2022pandemic}
D.~Tiwari, B.~S. Bhati, F.~Al-Turjman, and B.~Nagpal, ``Pandemic coronavirus
  disease (covid-19): World effects analysis and prediction using
  machine-learning techniques,'' \emph{Expert Systems}, vol.~39, no.~3, p.
  e12714, 2022.

\bibitem{raj2024classify}
S.~Raj, A.~Vishnoi, and A.~Srivastava, ``Classify alzheimer genes association
  using na{\"\i}ve bayes algorithm,'' \emph{Human Gene}, p. 201309, 2024.

\bibitem{ravi2023novel}
C.~Ravi, Y.~Yasmeen, K.~Masthan, R.~Tulasi, D.~Sriveni, and P.~Shajahan, ``A
  novel machine learning framework for tracing covid contact details by using
  time series locational data \& prediction techniques,'' \emph{International
  Journal on Recent and Innovation Trends in Computing and Communication},
  vol.~11, pp. 204--211, 2023.

\bibitem{gupta2021clustering}
M.~Gupta, R.~Kumar, S.~Chawla, S.~Mishra, and S.~Dhiman, ``Clustering based
  contact tracing analysis and prediction of sars-cov-2 infections,'' \emph{EAI
  Endorsed Transactions on Scalable Information Systems}, vol.~9, no.~35, 2021.

\bibitem{ravi2022critical}
J.~Ravi, S.~Kulkarni \emph{et~al.}, ``A critical review on density-based
  clustering algorithms and their performance in data mining,'' \emph{Int. J.
  Res. Anal. Rev.(IJRAR)}, vol.~9, no.~1, pp. 73--82, 2022.

\bibitem{masum2020covid}
A.~K.~M. Masum, S.~A. Khushbu, M.~Keya, S.~Abujar, and S.~A. Hossain,
  ``Covid-19 in bangladesh: a deeper outlook into the forecast with prediction
  of upcoming per day cases using time series,'' \emph{Procedia Computer
  Science}, vol. 178, pp. 291--300, 2020.

\bibitem{zhou2023improved}
L.~Zhou, C.~Zhao, N.~Liu, X.~Yao, and Z.~Cheng, ``Improved lstm-based deep
  learning model for covid-19 prediction using optimized approach,''
  \emph{Engineering applications of artificial intelligence}, vol. 122, p.
  106157, 2023.

\bibitem{rakhshan2023global}
S.~A. Rakhshan, M.~S. Nejad, M.~Zaj, and F.~H. Ghane, ``Global analysis and
  prediction scenario of infectious outbreaks by recurrent dynamic model and
  machine learning models: A case study on covid-19,'' \emph{Computers in
  Biology and Medicine}, vol. 158, p. 106817, 2023.

\bibitem{munoz2023new}
M.~Mu{\~n}oz-Organero, P.~Callejo, and M.~{\'A}. Hombrados-Herrera, ``A new rnn
  based machine learning model to forecast covid-19 incidence, enhanced by the
  use of mobility data from the bike-sharing service in madrid,''
  \emph{Heliyon}, vol.~9, no.~6, 2023.

\bibitem{vaswani2017attention}
A.~Vaswani, ``Attention is all you need,'' \emph{Advances in Neural Information
  Processing Systems}, 2017.

\bibitem{ming2023hostnet}
Z.~Ming, X.~Chen, S.~Wang, H.~Liu, Z.~Yuan, M.~Wu, and H.~Xia, ``Hostnet:
  improved sequence representation in deep neural networks for virus-host
  prediction,'' \emph{BMC bioinformatics}, vol.~24, no.~1, p. 455, 2023.

\bibitem{li2021long}
L.~Li, Y.~Jiang, and B.~Huang, ``Long-term prediction for temporal propagation
  of seasonal influenza using transformer-based model,'' \emph{Journal of
  biomedical informatics}, vol. 122, p. 103894, 2021.

\bibitem{wang2022predicting}
H.~Wang, G.~Tao, J.~Ma, S.~Jia, L.~Chi, H.~Yang, Z.~Zhao, and J.~Tao,
  ``Predicting the epidemics trend of covid-19 using epidemiological-based
  generative adversarial networks,'' \emph{IEEE Journal of Selected Topics in
  Signal Processing}, vol.~16, no.~2, pp. 276--288, 2022.

\bibitem{wang2023oriented}
Z.~Wang, P.~Zhang, Y.~Huang, G.~Chao, X.~Xie, and Y.~Fu, ``Oriented transformer
  for infectious disease case prediction,'' \emph{Applied Intelligence},
  vol.~53, no.~24, pp. 30\,097--30\,112, 2023.

\bibitem{Dataset}
``{Novel dataset},'' \url{Kaggle.com at
  https://www.kaggle.com/aayushiagrawall/novel-dataset}, 2024.

\bibitem{medhekar2013heart}
D.~S. Medhekar, M.~P. Bote, and S.~D. Deshmukh, ``Heart disease prediction
  system using naive bayes,'' \emph{Int. J. Enhanced Res. Sci. Technol. Eng},
  vol.~2, no.~3, 2013.

\bibitem{bengio1994learning}
Y.~Bengio, P.~Simard, and P.~Frasconi, ``Learning long-term dependencies with
  gradient descent is difficult,'' \emph{IEEE transactions on neural networks},
  vol.~5, no.~2, pp. 157--166, 1994.

\bibitem{gers2000learning}
F.~A. Gers, J.~Schmidhuber, and F.~Cummins, ``Learning to forget: Continual
  prediction with lstm,'' \emph{Neural computation}, vol.~12, no.~10, pp.
  2451--2471, 2000.

\bibitem{zhou2021informer}
H.~Zhou, S.~Zhang, J.~Peng, S.~Zhang, J.~Li, H.~Xiong, and W.~Zhang,
  ``Informer: Beyond efficient transformer for long sequence time-series
  forecasting,'' in \emph{Proceedings of the AAAI conference on artificial
  intelligence}, vol.~35, no.~12, 2021, pp. 11\,106--11\,115.

\bibitem{chakraborty2016extracting}
S.~Chakraborty and L.~Subramanian, ``Extracting signals from news streams for
  disease outbreak prediction,'' in \emph{2016 IEEE Global Conference on Signal
  and Information Processing (GlobalSIP)}.\hskip 1em plus 0.5em minus
  0.4em\relax IEEE, 2016, pp. 1300--1304.

\bibitem{kim2021infectious}
J.~Kim and I.~Ahn, ``Infectious disease outbreak prediction using media
  articles with machine learning models,'' \emph{Scientific reports}, vol.~11,
  no.~1, pp. 1--13, 2021.

\bibitem{bao2024robust}
J.~Bao, M.~Kudo, K.~Kimura, and L.~Sun, ``Robust embedding regression for
  semi-supervised learning,'' \emph{Pattern Recognition}, vol. 145, p. 109894,
  2024.

\bibitem{roy2023support}
A.~Roy and S.~Chakraborty, ``Support vector machine in structural reliability
  analysis: A review,'' \emph{Reliability Engineering \& System Safety}, vol.
  233, p. 109126, 2023.

\bibitem{kim2019weekly}
J.~Kim and I.~Ahn, ``Weekly ili patient ratio change prediction using news
  articles with support vector machine,'' \emph{BMC bioinformatics}, vol.~20,
  pp. 1--16, 2019.

\bibitem{thapen2016defender}
N.~Thapen, D.~Simmie, C.~Hankin, and J.~Gillard, ``Defender: detecting and
  forecasting epidemics using novel data-analytics for enhanced response,''
  \emph{PloS one}, vol.~11, no.~5, p. e0155417, 2016.

\bibitem{tanveer2024comprehensive}
M.~Tanveer, T.~Rajani, R.~Rastogi, Y.-H. Shao, and M.~Ganaie, ``Comprehensive
  review on twin support vector machines,'' \emph{Annals of Operations
  Research}, vol. 339, no.~3, pp. 1223--1268, 2024.

\bibitem{serban2019real}
O.~Șerban, N.~Thapen, B.~Maginnis, C.~Hankin, and V.~Foot, ``Real-time
  processing of social media with sentinel: A syndromic surveillance system
  incorporating deep learning for health classification,'' \emph{Information
  Processing \& Management}, vol.~56, no.~3, pp. 1166--1184, 2019.

\bibitem{shin2016high}
S.-Y. Shin, D.-W. Seo, J.~An, H.~Kwak, S.-H. Kim, J.~Gwack, and M.-W. Jo,
  ``High correlation of middle east respiratory syndrome spread with google
  search and twitter trends in korea,'' \emph{Scientific reports}, vol.~6,
  no.~1, p. 32920, 2016.

\bibitem{chae2018predicting}
S.~Chae, S.~Kwon, and D.~Lee, ``Predicting infectious disease using deep
  learning and big data,'' \emph{International journal of environmental
  research and public health}, vol.~15, no.~8, p. 1596, 2018.

\bibitem{drinkall2022forecasting}
F.~Drinkall, S.~Zohren, and J.~B. Pierrehumbert, ``Forecasting covid-19
  caseloads using unsupervised embedding clusters of social media posts,''
  \emph{arXiv preprint arXiv:2205.10408}, 2022.

\bibitem{sat-img}
I.~Lütkebohle, ``{THE IMPORTANCE OF SATELLITE IMAGES},''
  \url{https://www.geolandproject.eu/2022/04/14/the-importance-of-satellite-images/},
  2023, [Online; accessed 24-August-2023].

\bibitem{li2023geoimagenet}
W.~Li, S.~Wang, S.~T. Arundel, and C.-Y. Hsu, ``Geoimagenet: a multi-source
  natural feature benchmark dataset for geoai and supervised machine
  learning,'' \emph{GeoInformatica}, vol.~27, no.~3, pp. 619--640, 2023.

\bibitem{lee2024convolutional}
H.-J. Lee, S.-K. Mun, and M.~Chang, ``Convolutional lstm--lstm model for
  predicting the daily number of influenza patients in south korea using
  satellite images,'' \emph{Public Health}, vol. 230, pp. 122--127, 2024.

\bibitem{moukheiber2024multimodal}
D.~Moukheiber, D.~Restrepo, S.~A. Cajas, M.~P.~A. Montoya, L.~A. Celi, K.-T.
  Kuo, D.~M. L{\'o}pez, L.~Moukheiber, M.~Moukheiber, S.~Moukheiber
  \emph{et~al.}, ``A multimodal framework for extraction and fusion of
  satellite images and public health data,'' \emph{Scientific Data}, vol.~11,
  no.~1, p. 634, 2024.

\bibitem{yoneoka2024indirect}
D.~Yoneoka, A.~Eguchi, S.~Nomura, T.~Kawashima, Y.~Tanoue, M.~Hashizume, and
  M.~Suzuki, ``Indirect and direct effects of nighttime light on covid-19
  mortality using satellite image mapping approach,'' \emph{Scientific
  Reports}, vol.~14, no.~1, p. 25063, 2024.

\bibitem{peiffer2020machine}
N.~Peiffer-Smadja, S.~Delli{\`e}re, C.~Rodriguez, G.~Birgand, F.-X. Lescure,
  S.~Fourati, and E.~Rupp{\'e}, ``Machine learning in the clinical microbiology
  laboratory: has the time come for routine practice?'' \emph{Clinical
  Microbiology and Infection}, vol.~26, no.~10, pp. 1300--1309, 2020.

\bibitem{cabitza2021development}
F.~Cabitza, A.~Campagner, D.~Ferrari, C.~Di~Resta, D.~Ceriotti, E.~Sabetta,
  A.~Colombini, E.~De~Vecchi, G.~Banfi, M.~Locatelli \emph{et~al.},
  ``Development, evaluation, and validation of machine learning models for
  covid-19 detection based on routine blood tests,'' \emph{Clinical Chemistry
  and Laboratory Medicine (CCLM)}, vol.~59, no.~2, pp. 421--431, 2021.

\bibitem{alves2021explaining}
M.~A. Alves, G.~Z. Castro, B.~A.~S. Oliveira, L.~A. Ferreira, J.~A.
  Ram{\'\i}rez, R.~Silva, and F.~G. Guimar{\~a}es, ``Explaining machine
  learning based diagnosis of covid-19 from routine blood tests with decision
  trees and criteria graphs,'' \emph{Computers in Biology and Medicine}, vol.
  132, p. 104335, 2021.

\bibitem{RF}
``{What is random forest ?}'' \url{https://www.ibm.com/topics/random-forest},
  2024.

\bibitem{GRF}
``{A Comprehensive Guide to Random Forest: How It Works and Its
  Applications},''
  \url{https://graphite-note.com/a-comprehensive-guide-to-random-forest-how-it-works
  and-its applications/}, 2023.

\bibitem{brinati2020detection}
D.~Brinati, A.~Campagner, D.~Ferrari, M.~Locatelli, G.~Banfi, and F.~Cabitza,
  ``Detection of covid-19 infection from routine blood exams with machine
  learning: a feasibility study,'' \emph{Journal of medical systems}, vol.~44,
  no.~8, pp. 1--12, 2020.

\bibitem{banerjee2020use}
A.~Banerjee, S.~Ray, B.~Vorselaars, J.~Kitson, M.~Mamalakis, S.~Weeks,
  M.~Baker, and L.~S. Mackenzie, ``Use of machine learning and artificial
  intelligence to predict sars-cov-2 infection from full blood counts in a
  population,'' \emph{International immunopharmacology}, vol.~86, p. 106705,
  2020.

\bibitem{kukar2021covid}
M.~Kukar, G.~Gun{\v{c}}ar, T.~Vovko, S.~Podnar, P.~{\v{C}}ernel{\v{c}},
  M.~Brvar, M.~Zalaznik, M.~Notar, S.~Mo{\v{s}}kon, and M.~Notar, ``Covid-19
  diagnosis by routine blood tests using machine learning,'' \emph{Scientific
  reports}, vol.~11, no.~1, pp. 1--9, 2021.

\bibitem{yang2020routine}
H.~S. Yang, Y.~Hou, L.~V. Vasovic, P.~A. Steel, A.~Chadburn, S.~E.
  Racine-Brzostek, P.~Velu, M.~M. Cushing, M.~Loda, R.~Kaushal \emph{et~al.},
  ``Routine laboratory blood tests predict sars-cov-2 infection using machine
  learning,'' \emph{Clinical chemistry}, vol.~66, no.~11, pp. 1396--1404, 2020.

\bibitem{aljame2020ensemble}
M.~AlJame, I.~Ahmad, A.~Imtiaz, and A.~Mohammed, ``Ensemble learning model for
  diagnosing covid-19 from routine blood tests,'' \emph{Informatics in Medicine
  Unlocked}, vol.~21, p. 100449, 2020.

\bibitem{narmadha2020intelligent}
D.~Narmadha and A.~Pravin, ``An intelligent computer-aided approach for target
  protein prediction in infectious diseases,'' \emph{Soft Computing}, vol.~24,
  no.~19, pp. 14\,707--14\,720, 2020.

\bibitem{devi2018mso}
B.~R. Devi \emph{et~al.}, ``Mso--mlp diagnostic approach for detecting denv
  serotypes,'' \emph{International Journal of Pure and Applied Mathematics},
  vol. 118, no.~5, pp. 1--6, 2018.

\bibitem{MSO}
``{Multi-Swarm Optimization: A Powerful Approach to Solving Complex
  Problems},''
  \url{https://netinfo.click/AItools/lesson/?file=Multi-Swarm+Optimization&lang=en},
  2023.

\bibitem{zhou2023tempo}
B.~Zhou, H.~Zhou, X.~Zhang, X.~Xu, Y.~Chai, Z.~Zheng, A.~C. Kot, and Z.~Zhou,
  ``Tempo: A transformer-based mutation prediction framework for sars-cov-2
  evolution,'' \emph{Computers in Biology and Medicine}, vol. 152, p. 106264,
  2023.

\bibitem{ashraf2023early}
S.~Ashraf, M.~Kousar, and M.~S. Hameed, ``Early infectious diseases
  identification based on complex probabilistic hesitant fuzzy n-soft
  information,'' \emph{Soft Computing}, pp. 1--26, 2023.

\bibitem{zhu2022introduction}
X.~Zhu and A.~B. Goldberg, \emph{Introduction to semi-supervised
  learning}.\hskip 1em plus 0.5em minus 0.4em\relax Springer Nature, 2022.

\bibitem{sahoo2021potential}
P.~Sahoo, I.~Roy, R.~Ahlawat, S.~Irtiza, and L.~Khan, ``Potential diagnosis of
  covid-19 from chest x-ray and ct findings using semi-supervised learning,''
  \emph{Physical and Engineering Sciences in Medicine}, pp. 1--12, 2021.

\bibitem{sharma2018analysis}
N.~Sharma, V.~Jain, and A.~Mishra, ``An analysis of convolutional neural
  networks for image classification,'' \emph{Procedia computer science}, vol.
  132, pp. 377--384, 2018.

\bibitem{hussein2024auto}
A.~M. Hussein, A.~G. Sharifai, O.~M. Alia, L.~Abualigah, K.~H. Almotairi, S.~K.
  Abujayyab, and A.~H. Gandomi, ``Auto-detection of the coronavirus disease by
  using deep convolutional neural networks and x-ray photographs,''
  \emph{Scientific reports}, vol.~14, no.~1, p. 534, 2024.

\bibitem{issahaku2024multimodal}
F.-l.~Y. Issahaku, X.~Liu, K.~Lu, X.~Fang, S.~B. Danwana, and E.~Asimeng,
  ``Multimodal deep learning model for covid-19 detection,'' \emph{Biomedical
  Signal Processing and Control}, vol.~91, p. 105906, 2024.

\bibitem{hamida2021novel}
S.~Hamida, O.~El~Gannour, B.~Cherradi, A.~Raihani, H.~Moujahid, and H.~Ouajji,
  ``A novel covid-19 diagnosis support system using the stacking approach and
  transfer learning technique on chest x-ray images,'' \emph{Journal of
  Healthcare Engineering}, vol. 2021, no.~1, p. 9437538, 2021.

\bibitem{sadeghi2024potential}
A.~Sadeghi, M.~Sadeghi, A.~Sharifpour, M.~Fakhar, Z.~Zakariaei, M.~Sadeghi,
  M.~Rokni, A.~Zakariaei, E.~S. Banimostafavi, and F.~Hajati, ``Potential
  diagnostic application of a novel deep learning-based approach for
  covid-19,'' \emph{Scientific Reports}, vol.~14, no.~1, p. 280, 2024.

\bibitem{tan2024self}
Z.~Tan, Y.~Yu, J.~Meng, S.~Liu, and W.~Li, ``Self-supervised learning with
  self-distillation on covid-19 medical image classification,'' \emph{Computer
  Methods and Programs in Biomedicine}, vol. 243, p. 107876, 2024.

\bibitem{zhang1995prevention}
X.~Zhang, ``Prevention and control of emerging infectious diseases,'' 1995.

\bibitem{sood2021intelligent}
S.~K. Sood, V.~Sood, I.~Mahajan \emph{et~al.}, ``An intelligent healthcare
  system for predicting and preventing dengue virus infection,''
  \emph{Computing}, pp. 1--39, 2021.

\bibitem{valiakos2014use}
G.~Valiakos, K.~Papaspyropoulos, A.~Giannakopoulos, P.~Birtsas, S.~Tsiodras,
  M.~R. Hutchings, V.~Spyrou, D.~Pervanidou, L.~V. Athanasiou, N.~Papadopoulos
  \emph{et~al.}, ``Use of wild bird surveillance, human case data and gis
  spatial analysis for predicting spatial distributions of west nile virus in
  greece,'' \emph{PLoS One}, vol.~9, no.~5, p. e96935, 2014.

\end{thebibliography}


\end{document}